%% file: main.tex
\documentclass[twoside]{article}

\usepackage[accepted]{aistats2024}
\usepackage[round]{natbib}

\usepackage[utf8]{inputenc} 
\usepackage[T1]{fontenc}    
\usepackage{hyperref}       
\usepackage{url}            
\usepackage{booktabs}       
\usepackage{amsfonts}       
\usepackage{nicefrac}       
\usepackage{microtype}      
\usepackage{setspace}
\usepackage{amsmath}
\usepackage{amssymb, bm}
\usepackage[dvipsnames,svgnames]{xcolor}
\usepackage{xcolor}
\usepackage{hyperref}
\usepackage{mathrsfs}

\usepackage{amsthm}
\newtheorem{theorem}{Theorem}[section]
\newtheorem{lemma}[theorem]{Lemma}

\usepackage{algpseudocode}
\usepackage{algorithm}
\algrenewcommand\algorithmicindent{1em}%

\usepackage{graphicx}

\usepackage[round]{natbib}
\setcitestyle{authoryear,open={(},close={)}} 

\newcommand{\cmt}[1]{\textcolor{red}{\textbf{(AS: #1)}}}

\newcommand{\edit}[1]{\textcolor{red}{#1}}
\hypersetup{
	colorlinks=true,
	linkcolor=blue,
	urlcolor=red,
	citecolor=NavyBlue,
	linkbordercolor={0 0 1}
}
\begin{document}

%

%

\twocolumn[

\aistatstitle{$\alpha$-Fair Contextual Bandits}

\aistatsauthor{ Siddhant Chaudhary \And Abhishek Sinha }

\aistatsaddress{ Chennai Mathematical Institute \\ Chennai, India \And Tata Institute of Fundamental Research \\Mumbai 400005, India} ]

\input{abstract}
\input{related_work}
\input{full_information}
\input{full_information_olo}
\input{full_information_policy}
\input{full_information_policy_analysis}
\input{bandit}
\input{bandit_olo}
\input{bandit_policy}

\input{bandit_policy_analysis}
\input{putta_bound_proof}
\input{experiments}
\input{conclusion}
\clearpage
\bibliographystyle{unsrtnat}
\bibliography{bibmobility}
\input{appendix}
\end{document}

%% file: abstract.tex
\begin{abstract}
  Contextual bandit algorithms are at the core of many applications, including recommender systems, clinical trials, and optimal portfolio selection. One of the most popular problems studied in the contextual bandit literature is to maximize the sum of the rewards in each round by ensuring a sublinear regret against the best-fixed context-dependent policy. However, in many applications, the cumulative reward is not the right objective - the bandit algorithm must be fair in order to avoid the echo-chamber effect and comply with the regulatory requirements. In this paper, we consider the \textsc{$\alpha$-Fair Contextual Bandits} problem, where the objective is to maximize the global $\alpha$-fair utility function - a non-decreasing concave function of the cumulative rewards in the adversarial setting. The problem is challenging due to the non-separability of the objective across rounds. We design an efficient algorithm that guarantees an approximately sublinear regret in the full-information and bandit feedback settings.
\end{abstract}

%% file: related_work.tex
\section{Introduction and related work}
 In applications such as personalized recommendations, greedily optimizing for the most relevant content for each user profile tends to reduce the diversity of the recommended items as it induces an unhealthy \emph{echo-chamber} effect and propagates systematic biases \citep{celis2019controlling}.
Recall that standard contextual bandits with a separable cumulative utility function tend to maximize the click-through rates (CTR) by recommending the most popular item for each user profile \citep{semenov2022diversity}. However, an over-emphasis on the CTR metric invariably leads to polarization of opinions. A similar fairness issue arises with other popular recommender systems, such as movie or song recommendations by Netflix and Spotify and various online job recommendation portals. The main objective of this paper is to design a class of \emph{fair} contextual bandit algorithms equipped with a quantifiable fairness guarantee that holds even in the adversarial setting. Towards this goal, we propose a contextual bandit algorithm that maximizes the non-linear $\alpha$-fair utility function instead of the usual time-separable utility function. Due to the diminishing return property, the optimizer of the concave $\alpha$-fair utility function strikes a trade-off between the fairness and the accuracy of the recommendations through a tunable hyperparameter $\alpha \in [0,1).$ \cite{lan2010axiomatic} gave an axiomatic characterization of fair utility functions and showed that the $\alpha$-fair utility function comes out naturally. Other standard utility functions, \emph{e.g.,} proportional fair and min-max utilities, can be shown to be a limiting form of the $\alpha$-fair utility. 

Fairness in bandit and online convex optimization have been extensively studied in the literature  \citep{joseph2016fairness, chen2020fair, agarwal2014taming, patil2021achieving, si2022enabling, evandar2009, claure2020multi, li2019combinatorial}. \citet{chen2020fair} considered a fair contextual bandit problem with a finite number of contexts. Their online policy ensures that the probability of pulling each arm is lower-bounded by a pre-specified constant on every round. They establish a $O(\sqrt{TMN\log N})$ regret bound for the usual separable cumulative loss metric. In the stochastic setting, the work by \citet{patil2021achieving, claure2020multi}, and  \citet{li2019combinatorial} proposed constrained bandit policies that guarantee that the minimum \emph{fraction} of pulls of each arm exceeds a given threshold. Our work complements this line of work where we consider an unconstrained maximization of the non-separable $\alpha$-fair utility function. A detailed numerical comparison between our policy and the constrained bandit policy of \citet{chen2020fair} is presented in Section \ref{expts}. \citet{badanidiyuru2014resourceful} considered a similar contextual bandit problem in the stochastic setting, which was later extended to concave utility functions \citep{agrawal2014bandits, agrawal2016efficient}. \citet{agrawal2016efficient} gave an efficient policy with $O(\sqrt{T})$ regret in the stochastic setting. However, because of the impossibility of attaining a sublinear regret bound in the full-information setting \citep[Theorem 2]{sinha2023no}, their result can not be extended to the adversarial rewards, which is the main focus of this paper. Closest to this paper is the recent work by \citet{sinha2023no}, which considers the problem of maximizing the $\alpha$-fair utility function in the non-contextual full-information setting. In this paper, we extend their policy to the adversarial contextual bandit setting with finitely many contexts. This is accomplished by combining a recent scale-free bandit policy with non-separable rewards. 

\paragraph{Our contributions:} 
In this paper, we make the following contributions.
\begin{itemize}
    \item We propose an approximately no-regret contextual bandit algorithm for the $\alpha$-fair global utility function with an approximation factor at most $1.445$. The non-additivity of the $\alpha$-fair utility function across rounds makes this problem significantly more challenging than the classic contextual bandit problems.  We combine recent advances in online convex optimization and scale-free bandits to propose an efficient policy for this problem. 
    \item As a by-product of our algorithm specialized to a single context, we give the first fair MAB algorithm with an approximately sublinear regret for the $\alpha$-fair utility function in the adversarial setting.  
\item Because of the global non-separability of the utility function, we introduce a new analytical technique involving a novel \emph{bootstrapping} method to bound the regret in both full-information and bandit settings.
    \item We perform extensive numerical simulations of our policy and compare it with the state-of-the-art benchmarks with standard datasets.
\end{itemize}
All missing proofs can be found in the accompanying supplementary material.

%% file: full_information.tex
\section{The Full-information setting} \label{full-info}
We start our discourse with the simpler full-information setting where the entire reward vector for all arms is revealed to the policy at the end of every round. The more challenging bandit feedback setting, where only the reward component corresponding to the arm that was pulled is revealed on every round (where the event $3'$ takes place), will be studied in Section \ref{banditinfo}. Specifically, we consider a fully adversarial setting with $N$ arms \footnote{The arms could represent either distinct actions or $N$ different candidate policies for some problem from which we want to pick the best one \citep{auer2002nonstochastic}.} and a small number of contexts $M$. For structured contexts, one must reduce the number of distinct contexts, \emph{e.g.,} by clustering using similarity information \citep{slivkins2011contextual}, before using our algorithm. The following sequence of events takes place on every round $t \in [T]$.
\begin{enumerate}
    \item  The adversary first decides a context-reward pair $(c_t, r(t))$, where $c_t \in [M]$ and $\delta \leq r_i(t) \leq 1, \forall i\in [N].$  Here $\delta > 0$ is a fixed positive constant.
\item The context $c_t$ is revealed to the online policy, which then uses this information to choose an arm (possibly randomly) $I_t\in[N]$.
\item (\textbf{Full-Information Setting}) The policy obtains a reward of $r_{I_t}(t)$ and the entire reward vector $r(t)$ is revealed to the policy. Or,
\item[$3'.$] (\textbf{Bandit-feedback Setting}) The policy obtains a reward of $r_{I_t}(t)$ and only the value of $r_{I_t}(t)$ is revealed to the policy. 
\end{enumerate}
For a given online algorithm, let the probability vector $x^j(t) \in \Delta_N, j \in [M]$ denote the probability of pulling the arms when the $j$\textsuperscript{th} context is revealed to the policy on round $t$. An online policy is defined by the collection of (conditional) distributions $\big(x^j(t), j\in [M]\big),$ where, upon observing the current context $c_t$, the policy samples an arm $I_t\sim \bm{x}^{c_t}(t)$ for round $t$. 
The goal of the policy is to sequentially learn the best collection of distributions $\big(x^j(t), j \in [M]\big)$, one for each context, to maximize the $\alpha$-fair utility function described next. 


\subsubsection{Utility function and the regret metric}
For each arm $i\in[N]$, the (expected) cumulative reward accrued till round $t$ for a given policy is defined as:
\begin{align}
    \label{cumRewards}R_i(t) &= R_i(t - 1) + x^{c_t}_i(t)r_i(t), ~~ R_i(0)=1.
\end{align}
In this paper, we consider the problem of maximizing the sum of $\alpha$-fair utility functions of the arms where the utility of the $i$\textsuperscript{th} arm is defined as:
\begin{align}
    \label{utilityFunction}
    \phi(R_i(T)) := \dfrac{(R_i(T))^{1 - \alpha}}{1 - \alpha},~ i\in [N],
\end{align}
where $0\le \alpha < 1$ is some fixed constant. The parameter $\alpha$ strikes a trade-off between fairness and efficiency. Setting $\alpha=0$ corresponds to the usual linear reward function. On the other hand, larger $\alpha$ induces fairness because of the diminishing return property, which encourages playing all arms evenly \citep{lan2010axiomatic}. 
Formally, our objective is to design an online policy that minimizes the $c$-approximate \emph{contextual} regret, which competes with the best offline policy in hindsight (\emph{i.e.,} a fixed mapping from contexts to arms) instead of the best arm. Formally, the contextual regret is defined as:
\begin{eqnarray} \label{alphaFairRegretDefn}
    \text{Regret}_T(c) := {\max_{\bm{x_*}}}\sum_{i = 1}^N\phi(R_i^*(T)) - c\sum_{i = 1}^N\phi(R_i(T)),
\end{eqnarray}
where $c\ge 1$ is some small constant, and, for each user $i$, $R_i^*(T)$ is the cumulative reward \eqref{cumRewards} accrued by any static policy using the \textit{fixed} collection of distributions $\bm{x}_*\equiv (\bm{x}^1_*, ..., \bm{x}^M_*)$ used in Eq.\ \eqref{cumRewards}. A few words on the $c$-regret metric \eqref{alphaFairRegretDefn} are in order. Clearly, $c=1$ corresponds to the usual static regret. However, it is known from \citet[Theorem 2]{sinha2023no} that even in the full-information setting, no online policy can achieve a sublinear regret for $c =1.$ The concept of $c$-approximate regret has been useful in other online learning problems as well \citep{azar2022alpha, emamjomeh2021adversarial, paria2021texttt}.

\paragraph{Note:} 
1. We initialize $R_i(0)$ to  $1$ so that the derivative $\phi'(R_i(t))$ remains well-defined for all $t \in [T]$. 

2. In the full-information setting, we work exclusively with the expected cumulative rewards rather than the true rewards, which is stochastic due to the randomness of the policy. This allows us to carry out a simpler deterministic analysis. Using standard concentration inequalities, it can be shown that resulting bounds carry over for the true rewards as well \citep[Section 4]{sinha2023no}.
However, due to the limited feedback, this trick no longer works in the bandit setting, where we work with the stochastic true rewards. 

\begin{figure}[h]
    \centering
    \includegraphics[scale=0.22]{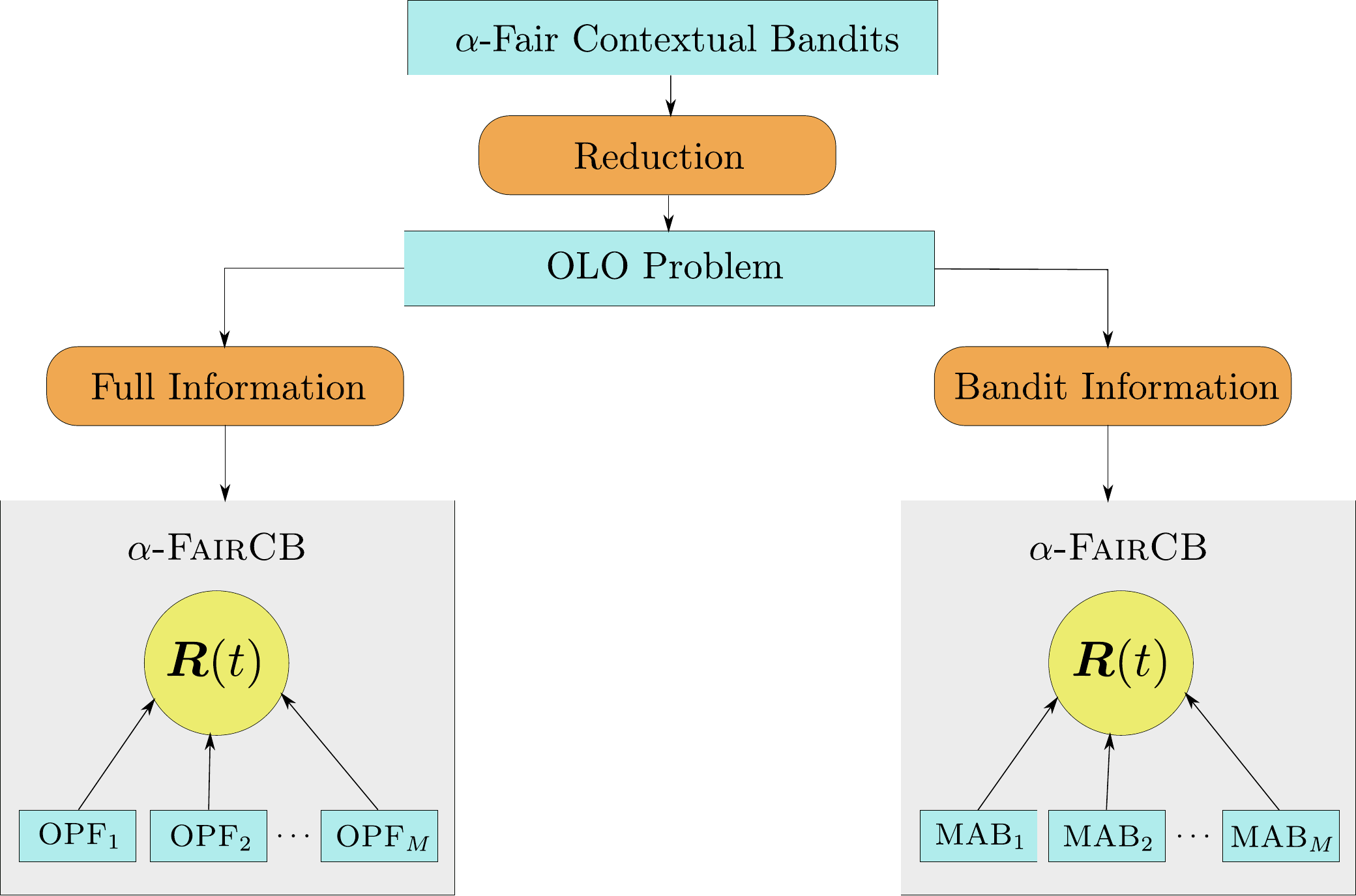}
    \caption{\scriptsize{Diagram representing the web of reductions used in the paper. First, the contextual bandit problem with a global $\alpha$-fair objective is reduced to a standard online linear optimization (OLO) problem. The reduction works the same way in both the full-information and bandit-information feedback settings. Then, in either setting, we parallelly run $M$ instances of a non-contextual policy, and all the $M$ policies are coupled through the \textit{shared} vector $\bm{R}(t)$ of cumulative rewards. On a high level, after the linearization step, the $j^{\text{th}}$ policy for $j\in[M]$ controls the regret for the $j$th context.}}
    \label{blockdiagram}
\end{figure}

%% file: full_information_olo.tex
\subsection{Algorithm design I: Linearization}

Similar to \citet{sinha2023no}, the algorithm design proceeds in two steps - (1) linearization with policy-dependent gradients and then (2) solving the linearized online optimization problem. See Figure \ref{blockdiagram} for a schematic. In the linearization step, we first reduce the problem to an instance of an online linear optimization (OLO) problem. 
Since the utility function $\phi(\cdot)$ is concave, we have  
\begin{align}
    \label{concavityProperty}
    \phi(x) - \phi(y)\le \phi'(y)(x - y)
\end{align}
for all $x, y > 0$. Now, let $\beta\ge 1$ be a constant, which will be fixed later. Taking $x = R_i^*(T)$ and $y = \beta R_i(T)$ in the above inequality, we get
\begin{align}
    &\phi(R_i^*(T)) - \beta^{1 - \alpha}\phi(R_i(T)) \nonumber\\ 
    &\overset{(a)}{=} \phi(R_i^*(T)) - \phi(\beta R_i(T)) \nonumber\\
    &\overset{(b)}{\le} \phi'(\beta R_i(T))[R_i^*(T) - \beta R_i(T)] \nonumber\\
    \label{oloReductionInquality}&\overset{(c)}{\le} \beta^{-\alpha} \phi'(R_i(T))\sum_{t = 1}^T r_i(t)[x^{c_t}_{*, i} - \beta x^{c_t}_i(t)],
\end{align}
where in $(a)$, we have used the property that $\phi(\beta x) = \beta^{1 - \alpha}(x)$ which holds for \eqref{utilityFunction}; in $(b)$, we have used inequality (\ref{concavityProperty}), and in $(c)$, we have used the definition of the cumulative rewards given in (\ref{cumRewards}), the fact that $\beta\ge 1$ and the property  $\phi'(\beta x) = \beta^{-\alpha}\phi'(x)$. Summing up the bound (\ref{oloReductionInquality}) over all the arms $i\in[N]$, we obtain the following bound to the $\beta^{1 - \alpha}$-approximate regret of any online policy:
\begin{align}
    \label{alphaFairRegretUB}
    &\text{Regret}_T(\beta^{1 - \alpha})\nonumber\\
    &\le \beta^{-\alpha}\sum_{t = 1}^T\sum_{i\in[N]}\phi'(R_i(T))r_i(t)[x^{c_t}_{*, i} - \beta x^{c_t}_i(t)].
\end{align}
Note that $R_i(T)$ is the cumulative reward accrued in the entire horizon of length $T$, and hence, it depends on the entire sequence of rewards and the actions of the policy. Clearly, this non-causal information is not available to the online policy at any intermediate round $t <T.$ This shows that directly minimizing the upper bound \eqref{alphaFairRegretUB} using online convex optimization methods is not feasible as the reward function involves the variables  $\phi'(R_i(T))$'s. 
To get around this fundamental difficulty, we now define 
a \textit{surrogate} online linear optimization problem by replacing the $t$\textsuperscript{th} coefficient $\phi'(R_i(T))$ in the RHS of the upper bound \eqref{alphaFairRegretUB} with its causal surrogate $\phi'(R_i(t - 1))$. With this substitution, the problem of minimizing \eqref{alphaFairRegretUB} becomes an instance of the online linear optimization (OLO) problem. However, in contrast with the standard OLO problem, here the reward functions are no longer oblivious as they depend on the policy through its past actions. By bounding the regret of the surrogate problem, we show that it is possible to derive an approximate regret bound to the original regret minimization problem \eqref{alphaFairRegretDefn}. Hence, dropping the factor $\beta^{-\alpha}$, the surrogate regret that we minimize is:
\begin{align}
    &\text{Surrogate Regret}_T \nonumber\\
    &= \max_{\bm{x}_*}\sum_{t = 1}^T\sum_{i\in[N]}\phi'(R_i(t - 1))r_i(t)[x^{c_t}_{*, i} - x^{c_t}_i(t)] \label{surrogateRegret}
\end{align}
In particular, for the surrogate problem, the linear reward vector at time step $t$ is given by $\phi'(\bm{R}(t - 1))\odot \bm{r}(t),$ which implicitly depends on the past actions of the policy (through the first term). Here, $\phi'(\bm{R}(t - 1)) \equiv (\phi'(R_1(t - 1)), ..., \phi'(R_N(t - 1)))$.
Upon setting $\beta\equiv (1-\alpha)^{-1},$ the following result relates the original regret (\ref{alphaFairRegretDefn}) with the surrogate regret (\ref{surrogateRegret}) for any policy.
\begin{lemma}
    \label{reductionBoundLemma}
    For any $T\ge 1$ and for any policy, we have
    \begin{align}
        \label{reductionBound}
        \emph{Regret}_T(c_\alpha) \le (1 - \alpha)^{\alpha} \emph{Surrogate Regret}_T + c_\alpha N
    \end{align}
    where $c_\alpha = (1 - \alpha)^{-(1 - \alpha)}\le e^{1/e} < 1.445$.
\end{lemma}
After accounting for $M$ different contexts with a common cumulative reward vector $\bm{R}(t)$, the proof generalizes the arguments in \citet[Lemma 1]{sinha2023no}. See Section \ref{reductionBoundLemma_proof} in the Appendix for the complete proof.

%% file: full_information_policy.tex
\subsection{Algorithm design II: Solving the linearized problem with full information}
In view of the regret bound (\ref{reductionBound}), we now propose $\alpha\textsc{-FairCB}$ - an online policy to approximately minimize the surrogate regret (\ref{surrogateRegret}). 
In brief, $\alpha\textsc{-FairCB}$ runs $M$ instances of adaptive online gradient descent policy in parallel, where the $j$\textsuperscript{th} instance is responsible for controlling the regret for the $j$\textsuperscript{th} context. These parallel policies are coupled through the common state vector $\bm{R}(t)$ - the cumulative reward accrued up to time $t$, which is affected by all contexts. Technically, this strategy works because, after the linearization step above, using the Cauchy-Scwarz inequality, the regret can be upper-bounded by the sum of policy-dependent gradients over all $M$ instances. Finally, the norm of these policy-dependent gradients are controlled using a novel \emph{bootstrapping} technique. The following lemma gives a precise regret bound for the surrogate problem.  

\begin{lemma}    \label{policySurrogateRegretBound}
    The $\alpha\textsc{-FairCB}$ policy described in \textbf{Algorithm \ref{fullInformationPolicy}} achieves the following static regret bound for the surrogate problem \emph{(\ref{surrogateRegret})}: 
    \begin{align}
        \emph{Surrogate Regret}_T = \begin{cases}
            O(N^3MT^{1/2 - \alpha}),&  \text{if }0 < \alpha < \frac{1}{2} \\
            O(N^3M\sqrt{\log T}),&  \text{if } \alpha = \frac{1}{2} \\
            O(1),&  \text{if }\frac{1}{2} < \alpha < 1.
        \end{cases}
    \end{align}
\end{lemma}
See Section \ref{sregretpf} for the proof of the result.
The proof of this lemma exploits a novel \emph{bootstrapping technique} which repeatedly boosts the estimate of the gradients, which are controlled by the policy, to obtain a better adaptive regret bound. 
Combining \textbf{Lemma} \ref{reductionBoundLemma} and \textbf{Lemma} (\ref{policySurrogateRegretBound}), we establish our main result.

\begin{theorem}
     \textbf{Algorithm \ref{fullInformationPolicy}} achieves the following approximate regret bound for the contextual bandit problem in the full information setting with the $\alpha$-fair utility function:
     \begin{align}
        \emph{Regret}_T(c_\alpha) = (1 - \alpha)^{\alpha} \begin{cases}
            O(N^3MT^{1/2 - \alpha}), \text{if }0 < \alpha < \frac{1}{2} \\
            O(N^3M\sqrt{\log T}),  \text{if } \alpha = \frac{1}{2} \\
            O(1),~ \text{if }\frac{1}{2} < \alpha < 1.
        \end{cases}\nonumber
    \end{align}
    where $c_\alpha = (1 - \alpha)^{-(1 - \alpha)}<1.445$.
\end{theorem}

\begin{algorithm}
    \caption{$\alpha\textsc{-FairCB}$ (Full Information Setting)}\label{fullInformationPolicy}
    \begin{algorithmic}[1]
        \State \textbf{Input:} Fairness parameter $0\le \alpha < 1$, Sequence of reward vectors $\bm{r}(1)$, ..., $\bm{r}(T)$, Sequence of contexts $c_1, ..., c_T$, Euclidean projection oracle on the simplex $\Pi_{\Delta_N}$, and an upper bound $D = \sqrt{2}$ to the Euclidean diameter of the simplex $\Delta_N$.

        \State \textbf{Output:} Distributions $\bm{x}^{c_t}(t)$ for each round $t$.

        \State \textbf{Initialization}:
        \begin{align*}
            R_i(0)\gets 1,
            S_j \gets 0, 
            \bm{x}^j \gets \dfrac{\bm{1}}{N}, ~\forall i,j.
        \end{align*}
        \For{$t = 1$ to $T$}
            \State Receive the context $c_t$ for round $t$.

            \If{Context $c_t$ is seen for the first time}
                \State Output $\bm{x}^{c_t}(t) = \bm{x}^{c_t}$ (uniform distribution).
            \Else
                \State Let $t'$ be the last time step when context $c_t$ was seen.
                
                \State Compute gradient vector $\bm{g}$ as follows: 
                    \begin{align*}
                        g_i = \dfrac{r_i(t')}{R_i^\alpha}\quad \forall i\in[N]
                    \end{align*}

                \State Update the cumulative gradient norm:
                    \begin{align*}
                        S_{c_t}\gets S_{c_t} + \lVert \bm{g}\rVert_2^2
                    \end{align*}

                \State Carry out the online gradient ascent update:
                    \begin{align*}
                        \bm{x}^{c_t}\gets \Pi_{\Delta_N}\left(\bm{x}^{c_t} + \dfrac{D}{\sqrt{2 S_{c_t}}}\bm{g}\right)
                    \end{align*}
    
                \State Output $\bm{x}^{c_t}(t) = \bm{x}^{c_t}$.        
            \EndIf

            \State Observe reward vector $\bm{r}(t)$.

            \State Update $R_i(t)\gets R_i(t - 1) + x^{c_t}_i(t)r_i(t)$.
        \EndFor
    \end{algorithmic}
\end{algorithm}

%% file: bandit.tex
\section{The Bandit feedback setting} \label{banditinfo}

We now study the same problem in the more challenging bandit feedback model. In this setup, only the reward of the arm selected by the policy, \emph{i.e.,} $r_{I_t}^{c_t}(t)$, is revealed on each round. Following standard practice, we assume that the reward vectors $\bm{r}(t)$ and the context sequence $c_t\in[M]$ for each time step $t$ are generated by an \textit{oblivious adversary}, \emph{i.e.}, the sequence of rewards and contexts is fixed \emph{a priori}.

Because of the limited feedback, an online policy cannot observe the expected cumulative rewards defined in Eqn.\ \eqref{cumRewards} as one needs to know the entire reward vector $\bm{r}(t)$ to compute the expected reward. Hence, instead of using the distribution $\bm{x}^{c_t}(t)$, we directly use the random \textit{one-hot encoded vector} $\bm{X}^{c_t}(t)$ to define the true cumulative rewards \footnote{With a slight abuse of notation, we use the same symbol $\bm{R}(t)$ to denote the expected cumulative rewards in the full-information setting \eqref{cumRewards} and true cumulative rewards in the bandit feedback setting \eqref{cumRewardsBandit}.}. Here, the $I_t$ \textsuperscript{th} component (which corresponds to the selected arm) of the vector $\bm{X}^{c_t}(t)$ is set to one, and the rest of the components are set to zero. Hence, the true cumulative reward vector, which the policy can observe under the bandit feedback setting, evolves as follows:
\begin{align}
    \label{cumRewardsBandit}R_i(t) &= R_i(t - 1) + X^{c_t}_i(t)r_i(t), ~R_i(0)=1.
\end{align}
As before, we will use the notation $\bm{x}^{c_t}(t)\in\Delta_N$ to denote the probability distribution of pulling the arms on step $t$. Hence, for all $i\in[N]$ and $t\in[1, T]$, we have
\begin{align}
    \label{oneHotVector}
    \mathbb{P}[X^{c_t}_i(t) = 1] = x^{c_t}_i(t). 
\end{align}

Our objective is to design a policy which minimizes the expected $c$-approximate regret defined below:
\begin{align}
    &\text{Regret}_T(c) \nonumber\\
    \label{contextualalphaPseudoRegret} &:= \max_{\bm{x}_*\in(\Delta_N)^M}\mathbb{E}\left[\sum_{i\in[N]}\phi(R_i^*(T)) - c\sum_{i\in[N]}\phi(R_i(T))\right].
\end{align}
In the above definition, $c\ge 1$ is a small constant whose value will be specified later and $\bm{R}^*(T)$ is the cumulative reward vector obtained for a stationary contextual bandit policy which pulls arms according to the fixed collection of distributions $\bm{x}_* \equiv (\bm{x}^1_*, ..., \bm{x}^M_*)$ depending on the current context. 
 Let $(\bm{x}^1_*, ..., \bm{x}^M_*)\in(\Delta_N)^M$ be the best-fixed collection of distributions which achieves the maximum in (\ref{contextualalphaPseudoRegret}). We have
\begin{align}
    \label{jensenOnRegretDefn}
    &\text{Regret}_T(c) \nonumber \\
    &=\mathbb{E}\left[\sum_{i\in[N]}\phi(R_i^*(T)) - c\sum_{i\in[N]}\phi(R_i(T))\right] \nonumber\\
    &\overset{(a)}{=} \sum_{i\in[N]}\mathbb{E}[\phi(R_i^*(T))   ] - c\mathbb{E}\left[\sum_{i\in[N]}\phi(R_i(T))\right] \nonumber\\
    &\overset{(b)}{\le} \sum_{i\in[N]}\phi(\mathbb{E}[R_i^*(T)]) - c\mathbb{E}\left[\sum_{i\in[N]}\phi(R_i(T))\right] \nonumber\\
    &\overset{(c)}{=} \sum_{i\in[N]}\phi\left(1 + \sum_{t = 1}^T r_i(t)x^{c_t}_{*, i}\right) - c\mathbb{E}\left[\sum_{i\in[N]}\phi(R_i(T))\right]
\end{align}
Above, in $(a)$, we have used the linearity of expectation. In $(b)$, we have used Jensen's Inequality on the concave function $\phi$. In $(c)$, we have just expanded $\mathbb{E}[R_i^*(T)]$ using (\ref{cumRewardsBandit}) and (\ref{oneHotVector}).

%% file: bandit_olo.tex
\subsection{Algorithm design I: Linearization }
Similar to the full-information setting, we handle the non-linearity by reducing the problem to a standard bandit problem with appropriately constructed linear reward functions. Following (\ref{oloReductionInquality}), we have
\begin{align}
    &\phi(\mathbb{E}R_i^*(T)) - \beta^{1 - \alpha}\phi(R_i(T)) \nonumber\\
    &\le \beta^{-\alpha}\phi'(R_i(T))\sum_{t = 1}^Tr_i(t)[x^{c_t}_{*, i} - \beta X^{c_t}_i(t)]
\end{align}
where above, $\beta\ge 1$ is some constant to be fixed later. Summing the above inequality for all $i\in[N]$ and taking expectations w.r.t the actions of the policy, we have 
\begin{align}
    &\sum_{i\in[N]}\phi(\mathbb{E}R_i^*(T)) - \beta^{1 - \alpha}\mathbb{E}\left[\sum_{i\in[N]}\phi(R_i(T))\right] \nonumber\\
    &\le \beta^{-\alpha}\mathbb{E}\left[\sum_{i\in[N]}\sum_{t = 1}^T\phi'(R_i(T))r_i(t)[x^{c_t}_{*, i} - \beta X^{c_t}_i(t)]\right].
\end{align}
Combining the last inequality with (\ref{jensenOnRegretDefn}), we get
\begin{align}
    &\text{Regret}_T(\beta^{1 - \alpha}) \nonumber\\
    \label{alphaFairRegretUBBandit}&\le \beta^{-\alpha}\mathbb{E}\left[\sum_{i\in[N]}\sum_{t = 1}^T\phi'(R_i(T))r_i(t)[x^{c_t}_{*, i} - \beta X^{c_t}_i(t)]\right].
\end{align}
Motivated by the above bound, we now consider a surrogate bandit problem by replacing the term $\phi'(R_i(T))$ with its causal counterpart $\phi'(R_i(t - 1))$. We now design an online policy to minimize the surrogate regret defined as follows:
\begin{align}
    &\text{Surrogate Regret}_T \nonumber\\
    &\equiv \mathbb{E}\left[\sum_{i\in[N]}\sum_{t = 1}^T\phi'(R_i(t - 1))r_i(t)[x^{c_t}_{*, i} - X^{c_t}_i(t)]\right] \nonumber\\
    \label{banditSurrogateRegret}&= \mathbb{E}\left[\sum_{t = 1}^T \langle \phi'(\bm{R}(t - 1))\odot \bm{r}(t), \bm{x}^{c_t}_* - \bm{X}^{c_t}(t)\rangle\right].
\end{align}
As before, $\phi'(\bm{R}(t - 1))\equiv (\phi'(R_1(t - 1)), ..., \phi'(R_N(t - 1)))$. Analogous to \textbf{Lemma} \ref{reductionBoundLemma}, we have the following result, which relates the regret defined in (\ref{contextualalphaPseudoRegret}) to the surrogate regret defined in (\ref{banditSurrogateRegret}).

\begin{lemma}
    \label{reductionBoundLemmaBandit}
    For any $T\ge 1$, we have
    \begin{align}
        \label{reductionBoundBandit}
        \emph{Regret}_T(c_\alpha)\le (1 - \alpha)^{\alpha}\emph{Surrogate Regret}_T + c_\alpha N,
    \end{align}
    where $c_\alpha = (1 - \alpha)^{-(1 - \alpha)}\le e^{1/e} < 1.445$.
\end{lemma}

%% file: bandit_policy.tex
\subsection{Algorithm design II: Solving the linearized problem with bandit feedback}

Lemma \ref{reductionBoundLemmaBandit} motivates us to design an online policy that minimizes the regret \eqref{banditSurrogateRegret} for the surrogate bandit problem. However, unlike the standard adversarial bandit problem, where the reward functions are fixed \emph{a priori} in an oblivious fashion, in this case, the rewards for each round $t$, defined as $\bm{g}_t \equiv  \phi' (\bm{R}(t - 1))\odot \bm{r}_t,$ \emph{depends on the past actions} of the policy. We can decompose the surrogate regret over the contexts as follows:
\begin{align}
    &\mathbb{E}\left[\sum_{t = 1}^T \langle \bm{g}_t, \bm{x}^{c_t}_* - \bm{X}^{c_t}(t)\rangle\right] \nonumber\\
    &= \mathbb{E}\left[\sum_{j\in[M]}\sum_{t:c_t = j}\langle \bm{g}_t, \bm{x}^j_* - \bm{X}^j(t)\rangle\right]\nonumber\\
    &\overset{(a)}{=} \sum_{j\in[M]} \mathbb{E}\left[\sum_{t:c_t = j}\langle \bm{g}_t, \bm{x}^j_* - \bm{X}^j(t)\rangle\right]\nonumber\\
    &\overset{(b)}{\le} \sum_{j\in[M]}\underbrace{\mathbb{E}\left[\max_{y\in\{\bm{e}_k\}_{k = 1}^N}\sum_{t:c_t = j} \langle \bm{g}_t, \bm{y} - \bm{X}^j(t)\rangle\right]}_{\textrm{regret for the}~j\textsuperscript{th}~ \textrm{context} } \nonumber\\
    \label{bestSurrogateRegretBandit}&=: \hat{\text{Regret}}_T.
\end{align}
Above, in $(a)$, we have used the linearity of expectation, in $(b)$, we have used the fact that for any fixed sequence of rewards in a bandit OLO problem, the best offline benchmark is the \textit{best-fixed arm in hindsight}. The above inequality can be written as
\begin{align}
    \label{bestSurrogateRegretInequalityBandit}
    \text{Surrogate Regret}_T\le \hat{\text{Regret}}_T
\end{align}
To minimize the surrogate regret, we now design a policy that minimizes $\hat{\text{Regret}}_T$, which is the sum of the regret for each context. Note that since the cumulative reward vector is common to all contexts, the regret bounds for different contexts are coupled with each other. To solve the per-context learning problem, we use the \textit{adaptive} and \textit{scale-free} multi-armed bandit policy, proposed by \citep{putta2022scale}, as a black box. Specifically, we run $M$ parallel instances of this policy, one for each context where they share the global cumulative reward vector $\bm{R}(t).$ For ease of reference, we quote regret bound achieved by the bandit policy of \citet{putta2022scale} in the following theorem.

\begin{algorithm}
    \caption{$\alpha\textsc{-FairCB}$ (Bandit Information Setting)}\label{banditInformationPolicy}
    \begin{algorithmic}[1]
        \State \textbf{Input:} Fairness parameter $0\le \alpha < 1$, Sequence of reward vectors $\bm{r}(1)$, ..., $\bm{r}(T)$, Sequence of contexts $c_1, ..., c_t$.

        \State \textbf{Output:} Arm $I_t\in[N]$ to be played at round $t$, for $t\in[1, T]$.

        \State Initialize $R_i(0)\gets 1$ for all $i\in[N]$.

        \State Initialize $M$ adaptive, scale-free MAB policies from \cite{putta2022scale}. Let $\mathscr{A}_j$ denote the $j$th instance of the policy, for $j\in[M]$.

        \For{$t = 1$ to $T$}
            \State Observe context $c_t$.

            \State Play an arm $I_t$ picked by policy $\mathscr{A}_{c_t}$. Let $\bm{X}^{c_t}(t)$ denote the one-hot vector representing arm $I_t$.

            \State Feed the modified reward vector $\phi'(\bm{R}(t - 1))\odot \bm{r}(t)$ to policy $\mathscr{A}_{c_t}$. \footnotemark

            \State Update $R_i(t)\gets R_i(t - 1) + X^{c_t}_i(t)r_i(t)$ for all $i\in[N]$.
        \EndFor
    \end{algorithmic}
    \label{Algorithm2}
\end{algorithm}
\footnotetext{Even though we pass the full vector $\phi'(\bm{R}(t - 1))\odot \bm{r}(t)$ to the bandit subroutine, it only ``sees'' the reward $\phi'(R_{I_t}(t - 1))r_{I_t}(t)$ for the arm $I_t$ it has just picked.}

\begin{theorem}[\textbf{Theorem 1} of \citep{putta2022scale}]
    \label{puttaOriginalBound}
    For any oblivious sequence of reward vectors $\bm{l}_1, ..., \bm{l}_T\in\mathbb{R}^N$, the adaptive version of \textbf{Algorithm 1} of \citet{putta2022scale} achieves the following regret bound:
    \begin{align} \label{putta-bd}
        \mathbb{E}\left[\max_{\{\bm{e}_k\}_{k = 1}^N}\sum_{t = 1}^T\langle \bm{l}_t, \bm{e}_k - \bm{X}(t)\rangle\right] \nonumber \\
        = O(\log T\cdot [\sqrt{NL_2} + L_\infty\sqrt{NL_1}]).
    \end{align}
   In the above, $\bm{X}(t)$ is the one-hot encoded vector denoting the arm pulled on round $t$, $L_\infty = \max_{t}\lVert \bm{l}_t\rVert_\infty$, $L_2 = \sum_{t = 1}^T\lVert \bm{l}_t\rVert_2^2$, $L_1 = \sum_{t = 1}^T\lVert \bm{l}_t\rVert_1$ and the expectation is taken w.r.t. the actions of the policy.
\end{theorem}
\paragraph{Remarks:} Technically, the regret bound in Theorem \ref{puttaOriginalBound} was originally established for oblivious adversaries. However, in our case, the surrogate reward vector $\bm{g}_t$ depends on the past actions of the policy up to round $t-1$. To see why we can still plug in the generic regret bound \eqref{putta-bd}, note that the reward vector $\bm{g}_t$ on round $t$ does not depend on the action $\bm{X}(t)$ taken on round $t$. Hence, we can use the regret bound for an \emph{imaginary} adversary that fixes the reward vector $\bm{g}_t$ at the end of round $t-1$. Since the reward on round $t$ does not affect the previous actions of the policy, the regret bound \eqref{putta-bd} holds. 
Adapting the above bound to our contextual setting, we have the following scale-free regret bound. 
\begin{lemma}
    \label{puttaRegretBound}
    For any $t\in[1, T]$, let $\bm{g}_t := \phi'(\bm{R}(t - 1))\odot \bm{r}(t)$. The adaptive version of \textbf{Algorithm 1} of \citep{putta2022scale} achieves the following bound for any $j\in[M]$:
    \begin{align}
        &\mathbb{E}\left[\max_{y\in\{\bm{e}_k\}_{k = 1}^N}\sum_{t:c_t = j} \langle \bm{g}_t, \bm{y} - \bm{X}^j(t)\rangle\right] \leq \nonumber\\
        &\tilde{O}\left(\mathbb{E}\left[\sqrt{N\sum_{t:c_t = j}\lVert\bm{g}_t\rVert_2^2} + \max_{t:c_t = j}\lVert \bm{g}_t\rVert_\infty\sqrt{N\sum_{t:c_t = j}\lVert\bm{g}_t\rVert_1}\right]\right),
    \end{align}
    where the $\tilde{O}(\cdot)$ notation hides the logarithmic factors. Above, the expectation is taken w.r.t the policy actions.
\end{lemma}
Please refer to Section \ref{puttaBoundProof} for the proof. The following result bounds the surrogate regret (\ref{bestSurrogateRegretBandit}).
\begin{lemma}
    \label{policySurrogateRegretBoundBandit}
    The $\alpha\textsc{-FairCB}$ policy described in \textbf{Algorithm} \ref{banditInformationPolicy} achieves the following bound on the regret of the surrogate bandit OLO problem for the $\alpha$-fair utility function:
    \begin{align}
        \hat{\emph{Regret}}_T = \tilde{O}(MN^2T^{\frac{1 - \alpha}{2}})
    \end{align}
    where the $\tilde{O}(\cdot)$ notation hides the $\log T$ factor.
\end{lemma}

Finally, combining \textbf{Lemma} \ref{reductionBoundLemmaBandit}, (\ref{bestSurrogateRegretInequalityBandit}) and \textbf{Lemma} \ref{policySurrogateRegretBoundBandit}, we establish our main result.

\begin{theorem}
    \textbf{Algorithm} \ref{banditInformationPolicy} achieves the following approximate regret bound for the contextual bandit problem in the bandit information feedback setting with the $\alpha$-fair utility function:
    \begin{align}
        \emph{Regret}_T(c_\alpha) = (1 - \alpha)^{\alpha}\tilde{O}(MN^2T^{\frac{1 - \alpha}{2}})
    \end{align}
    where $c_\alpha = (1 - \alpha)^{-(1 - \alpha)}< 1.445$, and the $\tilde{O}$ notation hides factors logarithmic in $T$.
\end{theorem}

%% file: experiments.tex
\section{Experiments} \label{expts}

We evaluate the performance of the proposed algorithm on a movie genre recommendation problem using the \textsc{MovieLens} 25M dataset \citep{movielens}. The dataset consists of 25 million data points, each consisting of a movie rating given by a user. For our experiments, we take a small sample comprising of the first $5,000$ data points. The underlying contextual bandit problem is formulated as follows: we interpret the users as contexts and movie genres as arms. In the selected sample, the number of contexts turns out to be $M=33$, and the number of arms featured is $N=19$. The dataset is sorted by the column containing the timestamps at which the ratings were reported, and this is taken to be the order of request arrivals. Since our policy requires a positive lower bound to the rewards, we take the minimum reward to be $0.2$ if the recommended genre doesn't fit the current movie and $1$ otherwise. In our experiments, we study both the full information and the bandit information settings.

\textbf{Performance metrics:} We define the \textit{$\alpha$-performance} of a policy at time stamp $t\in[1, T]$ in these experiments as the total $\alpha$-fair utility:
\begin{align}
    \label{alpha-perf}\text{$\alpha$-Performance}(t) := \sum_{i\in[N]} \phi(R_i(t)).
\end{align}
To measure fairness, we use the popular \textit{Jain's Fairness Index} \citep{jain1998quantitative}, calculated for the vector of cumulative rewards at the end of the time horizon. For any round $t\in [1, T]$, Jain's fairness index is defined as:
\begin{align}
    \label{jain-fairness}\text{Jain's Fairness Index} := \dfrac{(\sum_{i = 1}^N R_i(t))^2}{N\sum_{i = 1}^N R_i^2(t)}.
\end{align}
Jain's fairness index assumes a value between $0$ and $1$, where a value of $1$ is obtained when all components of the reward vector are the same (i.e., fully fair). In particular, if each arm receives an equal share of cumulative rewards, this index will be $1$. Throughout our experiments, we take $\alpha=0.9$ (i.e. a high level of fairness). We also plot the approximate contextual regret as defined in equations (\ref{alphaFairRegretDefn}) and (\ref{contextualalphaPseudoRegret}) for the full information and bandit feedback settings, respectively.

\textbf{Calculating the offline baseline metrics:} Note that the offline benchmarks in equations (\ref{alphaFairRegretDefn}) and (\ref{contextualalphaPseudoRegret}) required for computing the approximate regret involve computing the best offline collection of $M$ distributions maximizing the cumulative $\alpha$-fair utility function. Since $\phi(\cdot)$ is a concave function, this is a standard concave maximization problem over the convex domain $(\Delta_N)^M$. In our experiments, we use the \texttt{CVXPY} package for solving this problem \citep{diamond2016cvxpy}.


\begin{figure*}[!htb]
\minipage{0.32\textwidth}
  \includegraphics[width=0.8\linewidth]{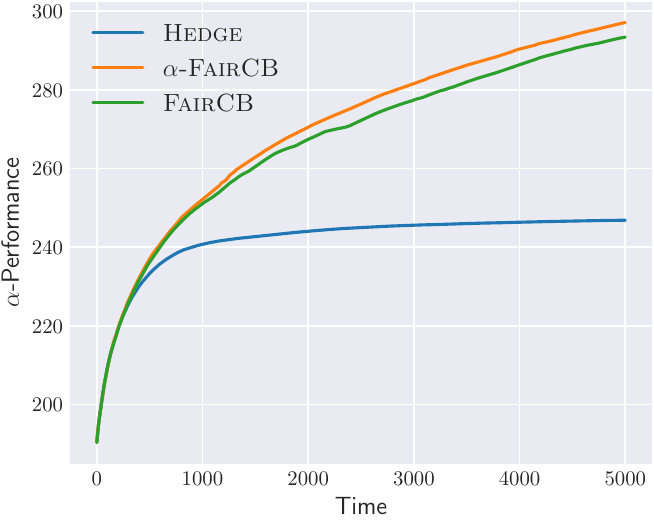}
  \caption{\scriptsize{$\alpha$-performance for the full information setting.}}\label{fig:performance_full_info}
\endminipage\hfill
\minipage{0.32\textwidth}
  \includegraphics[width=0.8\linewidth]{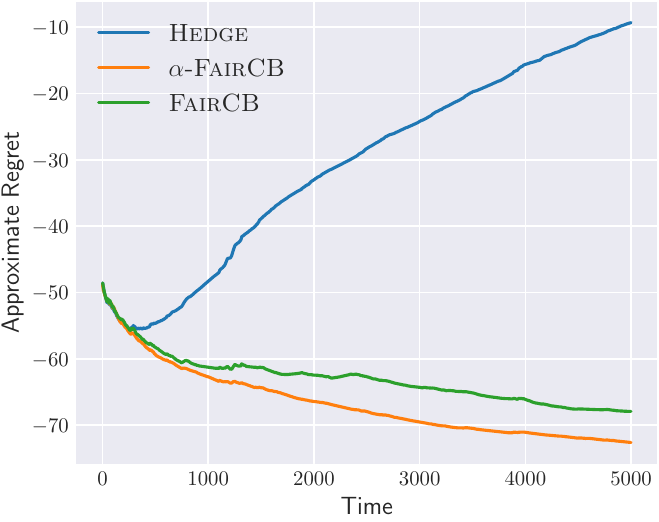}
  \caption{\scriptsize{Approximate regret for the full information setting.}}\label{fig:regret_full_info}
\endminipage\hfill
\minipage{0.32\textwidth}%
  \includegraphics[width=0.8\linewidth]{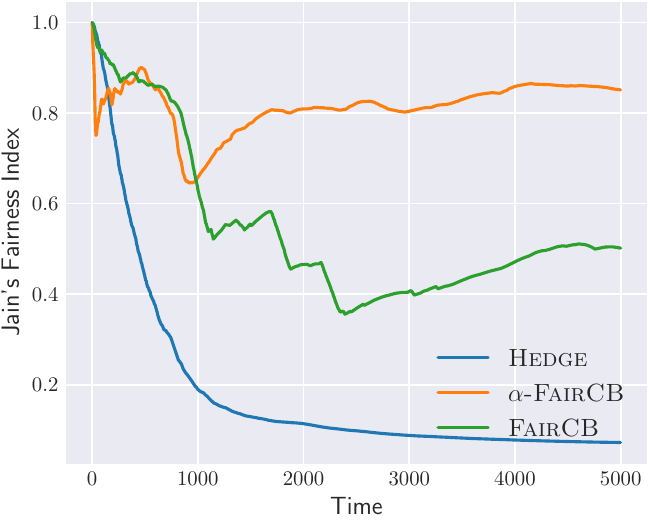}
  \caption{\scriptsize{Jain's Fairness Index for the full information setting.}}\label{fig:fairness_full_info}
\endminipage\hfill
\end{figure*}
\begin{figure*}[!htb]
\minipage{0.32\textwidth}
  \includegraphics[width=0.8\linewidth]{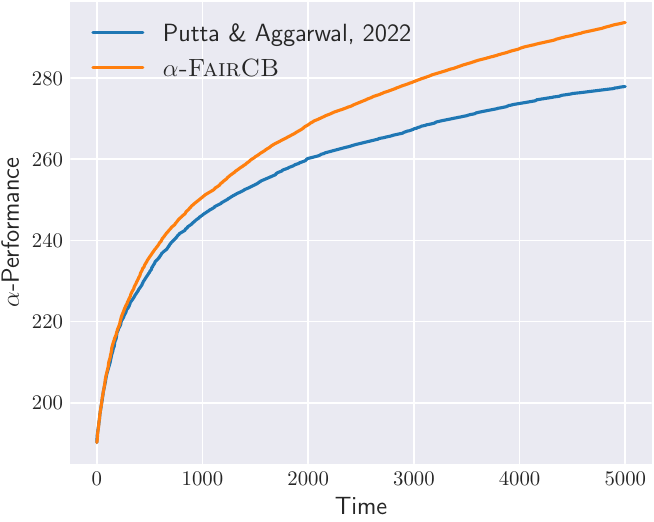}
  \caption{\scriptsize{$\alpha$-performance for the bandit information setting.}}\label{fig:performance_bandit_info}
\endminipage\hfill
\minipage{0.32\textwidth}
  \includegraphics[width=0.8\linewidth]{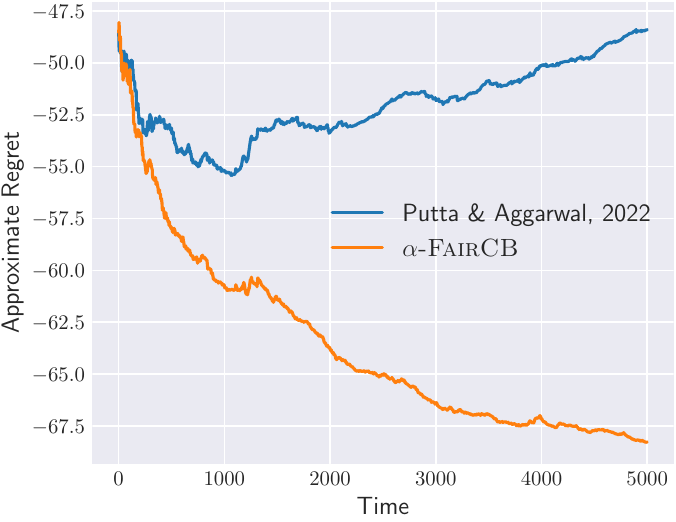}
  \caption{\scriptsize{Approximate regret for the bandit information setting.}}\label{fig:regret_bandit_info}
\endminipage\hfill
\minipage{0.32\textwidth}%
  \includegraphics[width=0.8\linewidth]{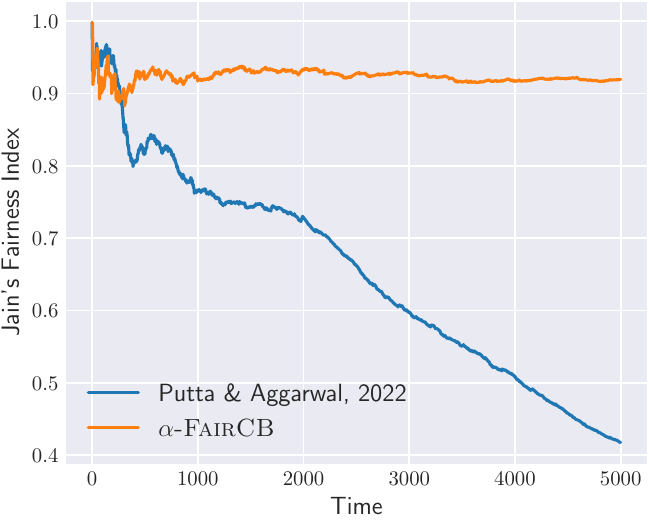}
  \caption{\scriptsize{Jain's Fairness Index for the bandit information setting.}}\label{fig:fairness_bandit_info}
\endminipage\hfill
\end{figure*}

\subsection{Experiments in the Full-information Setting}
\vspace{-5pt}

\textbf{Baseline Policies:} We consider two baselines (1) a context-agnostic \textsc{Hedge} policy (i.e. a policy that ignores contexts) and (2) the \textsc{FairCB} policy from \citep{chen2020fair}. Note that inherently \textsc{Hedge} is not a fair policy as its objective is to optimize the total reward. On the other hand, \textsc{FairCB}'s fairness constraint is specified by a tunable parameter $\nu\in(0, \frac{1}{N})$; in particular, the constraint is that the marginal probability of each arm being pulled at any given time step is at least $\nu$.  For our experiments, we consider $\nu = \frac{1}{2N}$ (note that $\frac{1}{N}$ is the largest possible fairness level that is allowed by \textsc{FairCB}). Note that the \textsc{FairCB} policy assumes the context distribution to be known; we simply generate this distribution offline by observing the sequence of contexts (users) in the dataset (and generating a distribution based on the frequencies of each user) and feed it back to the FairCB policy.

\textbf{Results:} Figure \ref{fig:performance_full_info} shows that the proposed $\alpha\textsc{-FairCB}$ policy outperforms the \textsc{Hedge} and \textsc{FairCB} policies in terms of $\alpha$-performance (\ref{alpha-perf}). As expected, the context-agnostic \textsc{Hedge} policy performs the worst among the three policies under consideration.  Consequently, $\alpha\textsc{-FairCB}$ achieves the lowest approximate regret among all the policies (Figure \ref{fig:regret_full_info}). Finally, in terms of Jain's Fairness Index (\ref{jain-fairness}), we observe that the proposed $\alpha\textsc{-FairCB}$ outperforms both the non-contextual \textsc{Hedge} and \textsc{FairCB} policies even for a moderately large time horizon (Figure \ref{fig:fairness_full_info}).

\subsection{Experiments in the Bandit Setting}

\textbf{Baseline Policies:} As a baseline policy, we run the context-agnostic \textit{adaptive} multi-armed bandit policy proposed by \citet{putta2022scale}, which is also used by our contextual bandit policy as a subroutine. 

\textbf{Results:} From Figure \ref{fig:performance_bandit_info}, it is observed that $\alpha\textsc{-FairCB}$ outperforms the policy by \citep{putta2022scale} in terms of $\alpha$-performance, and consequently $\alpha\textsc{-FairCB}$ achieves a lower approximate regret as well (as seen in Figure \ref{fig:regret_bandit_info}). In terms of Jain's Fairness Index, it is observed from Figure \ref{fig:fairness_bandit_info} that although for the first few rounds, \citep{putta2022scale}'s policy outperforms $\alpha\textsc{-FairCB}$, but over the entire time horizon, $\alpha\textsc{-FairCB}$ achieves a significantly better fairness index. The behaviour for the first few time steps can be explained by the fact that \citet{putta2022scale}'s policy has an exploration component, which makes the policy choose each arm with an approximately equal probability in the initial stages. However, since their policy maximizes the cumulative rewards, it achieves a worse fairness index over a longer horizon. See Section \ref{addl_expts} in the Appendix for additional experimental results.

%% file: conclusion.tex
\section{Conclusion and Future Work} \label{conclusion}
In this paper, we considered the problem of learning adversarial unstructured context-to-reward mapping and proposed an approximately regret-optimal policy in the full-information and bandit feedback setting. In the future, it will be interesting to design efficient algorithms for the case of structured contexts. Finally, similar to \citet{chen2020fair}, designing $\alpha$-fair bandit algorithms that guarantee a fixed fraction of pulls to each arm would also be interesting to investigate.

%% file: appendix.tex
\newpage
\onecolumn 
\section{Appendix} \label{appendix_section}
\subsection{Proof of Lemma \ref{reductionBoundLemma}} \label{reductionBoundLemma_proof}
Before proving the claim, we establish an auxiliary result that will be useful later.
\begin{lemma}
    Under any policy which updates the cumulative rewards of the $i$\textsuperscript{th} user $R_i(\cdot)$ as in \emph{(\ref{cumRewards})} $\forall i \in [N]$, the following inequality holds:
    \begin{align}
        \label{integralInequality}
        \phi'(R_i(t - 1))[R_i(t) - R_i(t - 1)]\le \int_{R_i(t - 1) - 1}^{R_i(t) - 1}\phi'(R)\emph{d}R.
    \end{align}
\end{lemma}
\begin{proof}
    Since $0\le \alpha < 1$, observe that the utility function $\phi(\cdot)$ given by Eq.\  (\ref{utilityFunction}) is well-defined on $[0, \infty)$ and is differentiable in $(0, \infty)$. Also, because $R_i(\cdot)$ is monotonically non-decreasing and $R_i(0) = 1$, we note that $R_i(t - 1) - 1\ge 0$ for all $t\in[1, T]$. By the fundamental theorem of calculus combined with the mean value theorem, we have
    \begin{align}
        \label{byftoc}
        \int_{R_i(t - 1) - 1}^{R_i(t) - 1}\phi'(R)\text{d}R = \phi'(c_0)[R_i(t) - R_i(t - 1)]
    \end{align}
    for some $c_0\in(R_i(t - 1) - 1, R_i(t) - 1)$; in particular, we have $c_0 < R_i(t) - 1$. Now, from the defintion (\ref{cumRewards}) observe that $R_i(t) - R_i(t - 1) = x^{c_t}_i(t)r_i(t)\le 1$, where we have used the fact that $x^{c_t}_i(t), r_i(t)\le 1$. This implies that $R_i(t) - 1\le R_i(t - 1)$, and hence, $c_0 < R_i(t - 1)$.

    Finally, since $\phi(\cdot)$ is concave, $\phi'(\cdot)$ is non-increasing; this implies that $\phi'(c_0)\ge \phi'(R_i(t - 1))$. Combining this with (\ref{byftoc}), the claim follows.
\end{proof}
We now establish Lemma \ref{reductionBoundLemma}.
\begin{proof}
    The upper bound for $\text{Regret}_T(\beta^{1 - \alpha})$ from Eq.\ \eqref{alphaFairRegretUB} can be split into the difference of two terms $A$ and $B$ as defined below:
    \begin{align}
        \label{splittedUB}
        &\text{Regret}_T(\beta^{1 - \alpha})\le \beta^{-\alpha}[A - \beta B],
    \end{align}
    where
    \begin{align}
        A &= \sum_{i\in[N]}\phi'(R_i(T))\sum_{t = 1}^Tr_i(t)x^{c_t}_{*, i},\\
        B &= \sum_{i\in[N]}\phi'(R_i(T))\sum_{t = 1}^Tr_i(t)x^{c_t}_{i}(t).
    \end{align}
    Also, let $A'$ and $B'$ denote the corresponding terms in the regret expression (\ref{surrogateRegret}) for the surrogate OLO problem. We will now bound the terms $A$ and $B$ in terms of $A'$ and $B'$, respectively.

    \textbf{Proving $A\le A'$:} Note that the utility function $\phi(\cdot)$ is concave, and hence its derivative is non-increasing. Also, from the recurrence equation for the cumulative rewards (\ref{cumRewards}), it is clear that under any policy, $R_i(\cdot)$ is non-decreasing for any $i\in[N]$. Hence, we see that $\phi'(R_i(t - 1))\ge \phi'(R_i(T))$ for all $t\in[1, T]$ and $i\in[N]$. This implies that 
    \begin{align}
        A &= \sum_{i\in[N]}\phi'(R_i(T))\sum_{t = 1}^Tr_i(t)x^{c_t}_{*, i} \nonumber\\
        &\le \sum_{i\in[N]}\phi'(R_i(t - 1))\sum_{t = 1}^Tr_i(t)x^{c_t}_{*, i} \nonumber\\
        &= A'
    \end{align}

    \textbf{Proving $B'\le (1 - \alpha)^{-1}(B + N)$:} We now argue that the following set of inequalities holds:
    \begin{align}
        B' &= \sum_{i}\sum_{t = 1}^T \phi'(R_i(t - 1))r_i(t)x^{c_t}_i(t)\nonumber\\
        &\overset{(a)}{=} \sum_{i}\sum_{t = 1}^T \phi'(R_i(t - 1))[R_i(t) - R_i(t - 1)]]\nonumber\\
        &\overset{(b)}{\le} \sum_{i}\sum_{t = 1}^T \int_{R_i(t - 1) - 1}^{R_i(t) - 1}\phi'(R)\text{d}R\nonumber\\
        &\overset{(c)}{\le} \sum_{i}\int_{0}^{R_i(T)} \phi'(R)\text{d}R\nonumber\\
        &\overset{(d)}{=} \sum_{i} \phi(R_i(T))\nonumber\\
        &\overset{(e)}{=} (1 - \alpha)^{-1}\sum_{i}\phi'(R_i(T))R_i(T)\nonumber\\
        \label{B'lessthanB}&\overset{(f)}{=} (1 - \alpha)^{-1}\sum_{i\in[N]}\phi'(R_i(T))\left(1 + \sum_{t = 1}^T x^{c_t}_i(t)r_i(t) \right)\nonumber\\
        &\overset{(h)}{\le} (1 - \alpha)^{-1}(B + N)
    \end{align}
    where in $(a),$ we have used the recurrence for $R_i(\cdot)$ as given in \eqref{cumRewards}. In $(b)$, we have used \eqref{integralInequality}. In $(c)$, we have simply used the fact that $R_i(0) - 1 = 0$ and $R_i(T) - 1\le R_i(T)$. In $(d)$, we have used the fundamental theorem of calculus and the fact that $\phi(0) = 0$. In $(e)$, we have used the fact that $x\phi'(x) = (1 - \alpha)\phi(x)$ which holds for the $\alpha$-fair utility function $\phi(\cdot)$. In $(f)$, we have used the definition of the cumulative rewards as in \eqref{cumRewards}. In $(h)$, we have used the definition of $B$ and the fact that $\phi'(x) = x^{-\alpha}\le 1$ for all $x\ge 1$.

    Now, the inequality $B'\le (1 - \alpha)^{-1}(B + N)$ implies that $(1 - \alpha)B' - N\le B$. Since $\beta > 0$, we have $\beta B\ge \beta(1 - \alpha)B' - \beta N$. Combining this with $A\le A'$, we have that
    \begin{align}
        A - \beta B\le A' - \beta(1 - \alpha)B' + \beta N.
    \end{align}
    Now, pick $\beta = (1 - \alpha)^{-1}$ (which ensures that $\beta \ge 1$), and hence we obtain
    \begin{align}
        A - \beta B\le A' - B' + (1 - \alpha)^{-1}N,
    \end{align}
    and from Eq.\ \eqref{splittedUB}, we see that
    \begin{align}
        \text{Regret}_T(c_\alpha)&\le (1 - \alpha)^{\alpha}(A' - B') + c_\alpha N\nonumber\\
        &= (1 - \alpha)^{\alpha}\text{Surrogate Regret}_T + c_\alpha N,
    \end{align}
    which completes the proof of the lemma.
\end{proof}
\subsection{Proof of Lemma \ref{policySurrogateRegretBound}}\label{sregretpf}

For ease of notation, let $(\bm{x}^{1}_*, ..., \bm{x}^{M}_*)\in(\Delta_N)^M$ be the collection of distributions achieving the maximum in equation (\ref{surrogateRegret}). Now, observe that $\text{Surrogate Regret}_T$ defined in (\ref{surrogateRegret}) for the surrogate problem can be split into the sum of regrets over each of the contexts as follows:
\begin{align}
    \text{Surrogate Regret}_T &=\sum_{t = 1}^T \langle \phi'(\bm{R}(t - 1))\odot \bm{r}(t), \bm{x}^{c_t}_* - \bm{x}^{c_t}(t)\rangle \nonumber\\
    &= \sum_{j\in[M]}\sum_{t:c_t = j} \langle \phi'(\bm{R}(t - 1))\odot \bm{r}(t), \bm{x}^{j}_* - \bm{x}^{j}(t)\rangle \nonumber\\
    \label{bestBenchmarkPerContext}&\overset{(a)}{\le} \sum_{j\in[M]} \underbrace{\max_{\bm{x}^j_\circ\in\Delta_N}\sum_{t:c_t = j} \langle \phi'(\bm{R}(t - 1))\odot \bm{r}(t), \bm{x}^{j}_\circ - \bm{x}^{j}(t)\rangle}_{\text{Regret for the $j^\text{th}$ context}}
\end{align}
where above in $(a)$, we have simply used that the regret w.r.t $\bm{x}^j_*$ for context $j$ is upper bounded by the regret associated to the \textit{best} offline benchmark $\bm{x}^j_
\circ$ for context $j$. 

Next, from the pseudocode of $\alpha\textsc{-FairCB}$ (Full Information Version, Algorithm \ref{fullInformationPolicy}), note that a Projected Online Gradient Ascent (OGA) policy with adaptive step sizes (\textbf{Theorem 4.14} of \citep{orabona2019modern}) controls the regret for each context $j\in[M]$. For the sake of completeness, we mention the complete statement of the regret guarantee of the OGA policy.
\begin{theorem}[Theorem 4.14 of \citep{orabona2019modern}]
    Let $\Delta\subset \mathbb{R}^d$ be a convex set with diameter $D$. Let us consider a sequence of linear reward functions with gradients $\{\bm{g}_t\}_{t\ge 1}$. Run the Online Gradient Ascent policy with step sizes $\eta_t = \dfrac{D}{\sqrt{2}\sum_{\tau = 1}^T\lVert\bm{g}_\tau\rVert^2}$, $1\le t\le T$. Then, the standard regret under the OGA policy can be upper bounded as follows:
    \begin{align}
       \label{OGABound} \text{Regret}_T\le D\sqrt{2\sum_{t = 1}^T\lVert\bm{g}_t\rVert^2}.
    \end{align}
\end{theorem}

Note that, for our case we have $D = \sqrt{2}$. So, by the regret bound of the OGA policy (\ref{OGABound}), for any $j\in[M]$ we have
\begin{align}
    &\max_{\bm{x}^j_\circ\in\Delta_N}\sum_{t:c_t = j}\langle \phi'(\bm{R}(t - 1))\odot \bm{r}(t), \bm{x}^j_\circ - \bm{x}^j(t)\rangle \nonumber\\
    &\le D\sqrt{2\sum_{t:c_t = j}\lVert \phi'(\bm{R}(t - 1))\odot \bm{r}(t)\rVert_2^2} \nonumber\\
    &\overset{(a)}{\le} D\sqrt{2\sum_{t:c_t = j}\lVert \phi'(\bm{R}(t - 1))\rVert_2^2} \nonumber\\
    \label{reciprocalSquareInequality}&\overset{(b)}{=} D\sqrt{2\sum_{t:c_t = j}\sum_{i\in[N]}\dfrac{1}{R_i^{2\alpha}(t - 1)}}
\end{align}
where above in $(a)$, we have used the fact that $\bm{r}(t)\le \bm{1}$ for all $t$, and in $(b)$ we have used the fact that $\phi'(x) = x^{-\alpha}$. Now, summing (\ref{reciprocalSquareInequality}) over all the contexts $j\in[M]$ and combining this with (\ref{bestBenchmarkPerContext}), we see that
\begin{align}
    \text{Surrogate Regret}_T &\le \sum_{j\in[M]}D\sqrt{2\sum_{t:c_t = j}\sum_{i\in[N]}\dfrac{1}{R_i^{2\alpha}(t - 1)}} \nonumber\\
    &= M\sum_{j\in[M]}\dfrac{1}{M}D\sqrt{2\sum_{t:c_t = j}\sum_{i\in[N]}\dfrac{1}{R_i^{2\alpha}(t - 1)}} \nonumber\\
    &\overset{(a)}{\le} DM\sqrt{\dfrac{2}{M}\sum_{j\in[M]}\sum_{t:c_t = j}\sum_{i\in[N]}\dfrac{1}{R_i^{2\alpha}(t - 1)}} \nonumber\\
    \label{mainBoundOLORegret} &= D\sqrt{M}\sqrt{2\sum_{t = 1}^T\sum_{i\in[N]}\dfrac{1}{R_i^{2\alpha}(t - 1)}},
\end{align}
where above in $(a)$, we have used Jensen's Inequality for the square root function. Using the fact that $R_i(t - 1)\ge 1$ for all $t$, bound (\ref{mainBoundOLORegret}) implies that 
\begin{equation} \label{sur-reg-init-bd}
    \text{Surrogate Regret}_T\le O(\sqrt{MNT}).
\end{equation}
In the following, we show that the above $O(\sqrt{T})$ regret bound can be substantially improved using a novel \emph{bootstrapping} technique described below. 

\paragraph{Bootstrapping:} Note that the adaptive regret bound depends on the sum of the norm of gradients of the reward vectors, which are controlled by the policy itself. This is in sharp contrast with the usual OCO setting where the policy does not explicitly control the gradients, and the final regret bound is given in terms of the sum of the norm of gradients as given in \eqref{OGABound}. The \emph{bootstrapping} technique starts with a trivial upper bound on the gradient norms and then uses the regret bound itself to improve the upper bounds on the gradient norms. This, in turn, improves the regret bound through the adaptive regret bound \eqref{OGABound}. The process is repeated a few times to get the best possible bound.

We now apply the general bootstrapping method to our problem.
Note that by the definition of $\text{Surrogate Regret}_T$ in (\ref{surrogateRegret}), we have the following inequality for any \textit{fixed} collection $(\bm{x}^1_0, ..., \bm{x}^M_0)\in (\Delta_N)^M$ of distributions:
\begin{align}
    \label{regretDefnInequality}
    &\sum_{t = 1}^T\langle \phi'(\bm{R}(t - 1))\odot \bm{r}(t), \bm{x}^{c_t}(t)\rangle \nonumber\\
    &\ge \sum_{t = 1}^T\langle \phi'(\bm{R}(t - 1))\odot \bm{r}(t), \bm{x}^{c_t}_0\rangle - \text{Surrogate Regret}_T.
\end{align}
Also, using the fact that $x^{c_t}_i(t)r_i(t) = R_i(t) - R_i(t - 1)$ and following the same calculations up to step (d) of (\ref{B'lessthanB}), we see that
\begin{align}
    \sum_{t = 1}^T\langle \phi'(\bm{R}(t - 1))\odot \bm{r}(t), \bm{x}^{c_t}(t)\rangle \le \sum_{i\in[N]}\phi(R_i(T)).
\end{align}
Combining the above inequality with (\ref{regretDefnInequality}), we have
\begin{align}
    &\sum_{i\in[N]}\phi(R_i(T))\nonumber\\
    \label{intermediateInequality}&\ge \sum_{t = 1}^T\langle \phi'(\bm{R}(t - 1))\odot \bm{r}(t), \bm{x}^{c_t}_0\rangle - \text{Surrogate Regret}_T.
\end{align}
Next, we lower bound $\phi'(\bm{R}(t - 1))$ by $\phi'(\bm{R}(T))$ and pick $\bm{x}^j_0 = \frac{1}{N}\bm{1}$ for all $j\in[M]$ (i.e., we pick the uniform distribution as an offline benchmark for each context). Doing so, and using the fact that $\bm{r}(t)\ge \delta\bm{1}$ for all $t$, we have
\begin{align}
    \sum_{t = 1}^T\langle \phi'(\bm{R}(t - 1))\odot \bm{r}(t), \bm{x}^{c_t}_0\rangle &\ge \sum_{t = 1}^T\langle \phi'(\bm{R}(T))\odot \bm{r}(t), \bm{x}^{c_t}_0\rangle \nonumber\\
    &= \sum_{i\in[N]}\sum_{t = 1}^T\phi'(R_i(T))r_i(t)\dfrac{1}{N} \nonumber\\
    &\ge T\sum_{i\in[N]}\phi'(R_i(T))\dfrac{\delta}{N}.
\end{align}
Plugging the last inequality in (\ref{intermediateInequality}), we conclude that
\begin{align}
    \sum_{i\in[N]}\phi(R_i(T))\ge T\sum_{i\in[N]}\phi'(R_i(T))\dfrac{\delta}{N} - \text{Surrogate Regret}_T.
\end{align}
Now, noting that $0 < R_i(T)\le T$ for all $i$, and that $\phi(\cdot)$ is monotone non-decreasing, we see that for any $i\in[N]$ the above inequality implies
\begin{align}
    \dfrac{NT^{1 - \alpha}}{1 - \alpha}\ge T\phi'(R_i(T))\dfrac{\delta}{N} - \text{Surrogate Regret}_T,
\end{align}
which implies the following inequality after dividing throughout by $T$ and replacing $\phi'(R_i(T))$ by $\frac{1}{R_i^\alpha(T)}$:
\begin{align}
    \dfrac{N}{(1 - \alpha)T^\alpha}\ge \dfrac{1}{R_i^\alpha(T)}\dfrac{\delta}{N} - \dfrac{\text{Surrogate Regret}_T}{T},
\end{align}
which is equivalent to
\begin{align}
    \label{cumRewardReciprocalInequality}
    \dfrac{1}{R_i^\alpha(T)}\le \dfrac{N}{\delta}\left[\dfrac{N}{(1 - \alpha)T^\alpha} + \dfrac{\text{Surrogate Regret}_T}{T}\right].
\end{align}
Now, from Eq.\ \eqref{sur-reg-init-bd}, we have the following preliminary bound $\text{Surrogate Regret}_T\le O(\sqrt{MNT})$. We use the \emph{bootstrapping} technique by plugging this in (\ref{cumRewardReciprocalInequality}) to derive the following improved bound on the cumulative reward accrued by the $i$\textsuperscript{th} arm.
\begin{align}
    \label{cumRewardReciprocalInequalityUpdated}
    \dfrac{1}{R_i^\alpha(T)}\le O\left(\dfrac{N^2\sqrt{MN}}{T^{\min(1/2, \alpha)}}\right),~ \forall i \in [N].
\end{align}
Now, we consider the following two cases:

\textbf{Case 1: $0\le \alpha\le 1/2$.} In this case, from (\ref{cumRewardReciprocalInequalityUpdated}) we see that $\frac{1}{R_i^\alpha(T)}\le O(\frac{N^2\sqrt{MN}}{T^\alpha})$, and hence $\frac{1}{R_i^{2\alpha}(T)}\le O(\frac{N^5M}{T^{2\alpha}})$. Note that this bound holds for all $T$. Hence, plugging this in (\ref{mainBoundOLORegret}), we get
\begin{align}
    \text{Surrogate Regret}_T\le O\left(DN^{\frac{5}{2}}M\sqrt{2\sum_{t = 2}^T \sum_{i\in[N]}\dfrac{1}{(t - 1)^{2\alpha}}}\right)
\end{align}
If $0\le \alpha < 1/2$, the above bound becomes $\text{Surrogate Regret}_T\le O\left(DN^3MT^{\frac{1}{2} - \alpha}\right)$. If $\alpha = \frac{1}{2}$, the above bound becomes $\text{Surrogate Regret}_T\le O\left(DN^3M\sqrt{\log T}\right)$.

\textbf{Case 2: $1/2 < \alpha < 1$.} In this case, bound (\ref{cumRewardReciprocalInequalityUpdated}) implies that $\frac{1}{R_i^\alpha(T)}\le \left(\dfrac{N^2\sqrt{MN}}{T^{1/2}}\right)$, and hence $\dfrac{1}{R_i^{2\alpha}(T)}\le O\left(\dfrac{N^5M}{T}\right)$. Again, this is true for all $T$. So, plugging this in (\ref{mainBoundOLORegret}), we get
\begin{align}
    \text{Surrogate Regret}_T&\le O\left(DN^{\frac{5}{2}}M\sqrt{2\sum_{t = 2}^T \sum_{i\in[N]}\dfrac{1}{(t - 1)}}\right) \nonumber\\
    &= O(DN^3M\sqrt{\log T})
\end{align}
Plugging this back in (\ref{cumRewardReciprocalInequality}), we get that $\frac{1}{R_i^\alpha(T)}\le O(\frac{N^5M}{T^{\alpha}})$, and hence $\frac{1}{R_i^{2\alpha}(T)}\le O(\frac{N^{10}M^2}{T^{2\alpha}})$. Again, note that this holds for all $T$. Hence, plugging this in (\ref{mainBoundOLORegret}), we see that
\begin{align}
    \text{Surrogate Regret}_T&\le O\left(DN^5M^{\frac{3}{2}}\sqrt{2\sum_{t = 2}^T \sum_{i\in[N]}\dfrac{1}{(t - 1)^{2\alpha}}}\right) \nonumber\\
    &= O(1)
\end{align}
where above, we have used the fact that $2\alpha> 1$.
\subsection{Proof of Lemma \ref{reductionBoundLemmaBandit}}
    The proof of \textbf{Lemma} \ref{reductionBoundLemma} works here with minor modifications. Again, the upper bound in (\ref{alphaFairRegretUBBandit}) for $\text{Regret}_T(\beta^{1 - \alpha})$ can be split into the difference of two terms $A$ and $B$ as follows:
    \begin{align}
        \label{splittedUBBandit}
        \text{Regret}_T(\beta^{1 - \alpha})\le \beta^{-\alpha}\mathbb{E}[A - \beta B]
    \end{align}
    where
    \begin{align}
        A &= \sum_{i\in[N]}\phi'(R_i(T))\sum_{t = 1}^Tr_i(t)x^{c_t}_{*, i}\\
        B &= \sum_{i\in[N]}\phi'(R_i(T))\sum_{t = 1}^Tr_i(t)X^{c_t}_{i}(t)
    \end{align}
    Also, let $A'$ and $B'$ denote the corresponding terms in the surrogate regret for the OLO problem defined in (\ref{banditSurrogateRegret}). Following the same argument as in the proof of \textbf{Lemma} \ref{reductionBoundLemma}, we can obtain $A\le A'$ and $B'\le (1 - \alpha)^{-1}(B + N)$.

    As before, the inequality $B'\le (1 - \alpha)^{-1}(B + N)$ implies that $(1 - \alpha)B' - N\le B$. Since $\beta > 0$, we have $\beta B\ge \beta(1 - \alpha)B' - \beta N$. Combining this with $A\le A'$, we see that
    \begin{align}
        A - \beta B\le A' - \beta(1 - \alpha)B' + \beta N.
    \end{align}
    Now, pick $\beta = (1 - \alpha)^{-1}$ (ensuring that $\beta \ge 1$), and hence, we obtain
    \begin{align}
        A - \beta B\le A' - B' + (1 - \alpha)^{-1}N.
    \end{align}
    Taking expectations w.r.t the policy actions, we get
    \begin{align}
        \mathbb{E}[A - \beta B]\le \mathbb{E}[A' - B'] + (1 - \alpha)^{-1}N.
    \end{align}
    Finally, from (\ref{splittedUBBandit}), we get
    \begin{align}
        \text{Regret}(c_\alpha)&\le \beta^{-\alpha}\mathbb{E}[A' - B'] + \beta^{-\alpha}(1 - \alpha)^{-1}N \nonumber\\
        &= (1 - \alpha)^{\alpha}\text{Surrogate Regret}_T + c_\alpha N,
    \end{align}
    completing the proof of the lemma.
\subsection{Proof of \textbf{Lemma} \ref{policySurrogateRegretBoundBandit}}

Consider some context $j\in[M]$. As before, for any $t\in[1, T]$ let $\bm{g}_t := \phi'(\bm{R}(t - 1))\odot \bm{r}(t)$. Then, we have the following set of inequalities considering the adaptive regret bound of the MAB policy handling context $j$:
\begin{align}
    &\tilde{O}\left(\mathbb{E}\left[\sqrt{N\sum_{t:c_t = j}\lVert\bm{g}_t\rVert_2^2} + \max_{t:c_t = j}\lVert \bm{g}_t\rVert_\infty\sqrt{N\sum_{t:c_t = j}\lVert\bm{g}_t\rVert_1}\right]\right) \nonumber\\
    &\overset{(a)}{\le}\tilde{O}\left(\mathbb{E}\left[\sqrt{N\sum_{t:c_t = j}\lVert\bm{g}_t\rVert_2^2} + \sqrt{N\sum_{t:c_t = j}\lVert\bm{g}_t\rVert_1}\right]\right) \nonumber\\
    &\overset{(b)}{\le}\tilde{O}\left(\mathbb{E}\left[\sqrt{N\sum_{t:c_t = j}\lVert\phi'(\bm{R}(t - 1))\rVert_2^2} + \sqrt{N\sum_{t:c_t = j}\lVert\phi'(\bm{R}(t - 1))\rVert_1}\right]\right) \nonumber\\
    &\overset{(c)}{=}\tilde{O}\left(\mathbb{E}\left[\sqrt{N\sum_{t:c_t = j}\sum_{i\in[N]}\dfrac{1}{R_i^{2\alpha}(t - 1)}} + \sqrt{N\sum_{t:c_t = j}\sum_{i\in[N]}\dfrac{1}{R_i^{\alpha}(t - 1)}}\right]\right)\nonumber\\
    &\overset{(d)}{\le}\tilde{O}\left(\mathbb{E}\left[\sqrt{N\sum_{t:c_t = j}\sum_{i\in[N]}\dfrac{1}{R_i^{\alpha}(t - 1)}}\right]\right)\nonumber\\
    &\overset{(e)}{\le}\tilde{O}\left(\sqrt{N\sum_{t:c_t = j}\sum_{i\in[N]}\mathbb{E}\dfrac{1}{R_i^{\alpha}(t - 1)}}\right).
\end{align}
Above, in $(a)$ we have used the fact that $\max_{t:c_t = j}\lVert \bm{g}_t\rVert_\infty\le 1$, which follows because $\bm{r}(t)\le \bm{1}$ and $\phi'(\bm{R}(t - 1))\le \bm{1}$. In $(b)$, we have used the fact that $\bm{r}(t)\le \bm{1}$. In $(c)$, we have used $\phi'(x) = x^{-\alpha}$. In $(d)$, we have used the fact that for each $i\in[N]$, $R_i(t - 1)\ge 1$. Finally, in $(e)$, we have applied Jensen's Inequality to the concave square root function. So, from the last inequality and \textbf{Lemma} \ref{puttaRegretBound}, we get
\begin{align}
    &\mathbb{E}\left[\max_{y\in\{\bm{e}_k\}_{k = 1}^N}\sum_{t:c_t = j} \langle \phi'(\bm{R}(t - 1)\odot \bm{r}(t)), \bm{y} - \bm{X}^j(t)\rangle\right] \nonumber\\
    &\le \tilde{O}\left(2\sqrt{N\sum_{t:c_t = j}\sum_{i\in[N]}\mathbb{E}\dfrac{1}{R_i^{\alpha}(t - 1)}}\right)
\end{align}
Summing the above inquality over all contexts $j\in[M]$, we get the following inequality on $\hat{\text{Regret}}_T$ defined in (\ref{bestSurrogateRegretBandit}):
\begin{align}
    \hat{\text{Regret}}_T &\le \sum_{j\in[M]}\tilde{O}\left(\sqrt{N\sum_{t:c_t = j}\sum_{i\in[N]}\mathbb{E}\dfrac{1}{R_i^{\alpha}(t - 1)}}\right) \nonumber\\
    &= M\sum_{j\in[M]}\dfrac{1}{M}\tilde{O}\left(\sqrt{N\sum_{t:c_t = j}\sum_{i\in[N]}\mathbb{E}\dfrac{1}{R_i^{\alpha}(t - 1)}}\right) \nonumber\\
    &\overset{(a)}{\le} M\tilde{O}\left(\sqrt{\sum_{j\in[M]}\dfrac{N}{M}\sum_{t:c_t = j}\sum_{i\in[N]}\mathbb{E}\dfrac{1}{R_i^{\alpha}(t - 1)}}\right) \nonumber\\
    \label{reciprocalSquareInequalityBandit}&= \sqrt{MN}\tilde{O}\left(\sqrt{\sum_{t = 1}^T\sum_{i\in[N]}\mathbb{E}\dfrac{1}{R_i^{\alpha}(t - 1)}}\right),
\end{align}
where in $(a)$ above, we have used Jensen's Inequality on the concave square root function. Note that this bound is similar to the bound in (\ref{mainBoundOLORegret}) for the full information feedback setting, with the only difference being in the exponent of the cumulative reward sequence ($2\alpha$ versus $\alpha$).

Next, we will derive a bound similar to (\ref{cumRewardReciprocalInequality}). Note that by the definition of $\hat{\text{Regret}_T}$ in (\ref{bestSurrogateRegretBandit}), we have the following inequality for any fixed collection of distributions $(\bm{x}^1_0, ..., \bm{x}^M_0)$:
\begin{align}
    &\sum_{j\in[M]} \mathbb{E}\left[\sum_{t:c_t = j}\langle \bm{g}_t, \bm{X}^{j}(t)\rangle\right] \nonumber\\
    &\ge \sum_{j\in[M]} \mathbb{E}\left[\sum_{t:c_t = j}\langle \bm{g}_t, \bm{x}^{j}_0\rangle\right] - \hat{\text{Regret}}_T.
\end{align}
Using the linearity of expectation, the above inequality can be written as
\begin{align}
    \label{regretDefnInequalityBandit}
    \mathbb{E}\left[\sum_{t = 1}^T\langle \bm{g}_t, \bm{X}^{c_t}(t)\rangle\right] \ge \mathbb{E}\left[\sum_{t = 1}^T\langle \bm{g}_t, \bm{x}^{c_t}_0\rangle\right] - \hat{\text{Regret}}_T.
\end{align}
Next, observing that $X^{c_t}_i(t)r_i(t) = R_i(t) - R_i(t - 1)$ and following the same calculations up to step (d) of (\ref{B'lessthanB}) and taking expectations w.r.t the policy actions, we get
\begin{align}
    \label{expectedRewardUpperBoundBandit}
    \mathbb{E}\left[\sum_{t = 1}^T\langle \bm{g}_t, \bm{X}^{c_t}(t)\rangle\right] \le \mathbb{E}\left[\sum_{i\in[N]}\phi(R_i(T))\right]
\end{align}
Lower bounding $\phi'(R_i(t - 1))$ by $\phi'(R_i(T))$ yields
\begin{align}
    \sum_{t = 1}^T \langle \bm{g}_t, \bm{x}_0^{c_t}\rangle\ge \sum_{t = 1}^T \langle \phi'(\bm{R}(T))\odot \bm{r}(t), \bm{x}_0^{c_t}\rangle.
\end{align}
Finally, taking expectations w.r.t. the policy actions, we get
\begin{align}
    \mathbb{E}\left[\sum_{t = 1}^T \langle \bm{g}_t, \bm{x}_0^{c_t}\rangle\right]\ge \mathbb{E}\left[\sum_{t = 1}^T \langle \phi'(\bm{R}(T))\odot \bm{r}(t), \bm{x}_0^{c_t}\rangle\right].
\end{align}
Now, let us take the offline benchmark policy to be the uniform distribution for all contexts, i.e., $\bm{x}^j_0 = \frac{1}{N}\bm{1}$ for all $j\in[M]$, which will imply that $\sum_{t = 1}^T r_i(t)x^{c_t}_{0, i}\ge \frac{\delta T}{N}$ for all $i\in[N]$. So, the RHS in the last equation can be lower bounded by $\sum_{i\in[N]}\mathbb{E}[\phi'(R_i(T))]\cdot \frac{\delta T}{N}$. Hence, we get
\begin{align}
    \mathbb{E}\left[\sum_{t = 1}^T \langle \bm{g}_t, \bm{x}_0^{c_t}\rangle\right]\ge \sum_{i\in[N]}\mathbb{E}[\phi'(R_i(T))]\cdot \dfrac{\delta T}{N}
\end{align}
So, from the last equation and equations (\ref{regretDefnInequalityBandit}) and (\ref{expectedRewardUpperBoundBandit}), we get
\begin{align}
    \mathbb{E}\left[\sum_{i\in[N]}\phi(R_i(T))\right]\ge \sum_{i\in[N]}\mathbb{E}[\phi'(R_i(T))]\cdot \dfrac{\delta T}{N} - \hat{\text{Regret}}_T
\end{align}
So from here, following the same steps as in the full information feedback setting, we obtain
\begin{align}
    \label{cumRewardReciprocalInequalityBandit}
    \mathbb{E}\left[\dfrac{1}{R_i^\alpha(T)}\right]\le \dfrac{N}{\delta}\left[\dfrac{N}{(1 - \alpha)T^\alpha} + \dfrac{\hat{\text{Regret}}_T}{T}\right]
\end{align}
Note the similarity between the above inequality and inequality (\ref{cumRewardReciprocalInequality}) for the full information setting.

Now, we know that $R_i^\alpha(t - 1)\ge 1$ for all $i\in[N]$ and $t$. Plugging this in (\ref{reciprocalSquareInequalityBandit}), we get our first bound, which is $\hat{\text{Regret}}_T\le \tilde{O}(N\sqrt{MT})$. 

As before, we do a tighter analysis to get a better regret bound. So, let $\alpha_0\in[0, 1)$ be any number. As the result of \textbf{Lemma} \ref{policySurrogateRegretBoundBandit} claims, we want to show that $\hat{\text{Regret}}_T = \tilde{O}(T^{\frac{1 - \alpha_0}{2}})$. Since $\alpha_0\in[0, 1)$, there is some positive integer $N_0\ge 0$ such that
\begin{align}
    \label{alphaKnotChoice}
    \dfrac{2^{N_0} - 1}{2^{N_0}}\le \alpha_0 < \dfrac{2^{N_0 + 1} - 1}{2^{N_0 + 1}}
\end{align}
Now, let $\epsilon_0 > 0$ be a very small number which satisfies the following inequalities for all $0\le n \le N_0$:
\begin{align}
    \label{epsilonIneq1}\dfrac{2^n - 1}{2^n} &< \dfrac{2^{n + 1} - 1}{2^{n + 1}} - \left(\dfrac{2^{n + 1} - 1}{2^n}\right)\epsilon_0\\
    \label{epsilonIneq2}\alpha_0 &\le \dfrac{2^{N_0 + 1} - 1}{2^{N_0 + 1}} - \left(\dfrac{2^{N_0 + 1} - 1}{2^{N_0}}\right)\epsilon_0
\end{align}
Note that, the above two conditions are equivalent to the following two conditions for all $0\le n\le N_0$:
\begin{align}
    \epsilon_0 &<\dfrac{2^{n}}{2^{n + 1} - 1}\left[\dfrac{2^{n + 1} - 1}{2^{n + 1}} - \dfrac{2^{n} - 1}{2^n}\right]\quad\\
    \epsilon_0 &<\dfrac{2^{N_0}}{2^{N_0 + 1} - 1}\left[\dfrac{2^{N_0 + 1} - 1}{2^{N_0 + 1}} - \alpha_0\right]
\end{align}
Since all the quantities on the RHS in the two equations above are positive, $\epsilon_0$ can be taken to be something smaller than the minimum of all the above quantities. Now, we have obtained $\hat{\text{Regret}}_T\le \tilde{O}(N\sqrt{MT}) = O(\log T\cdot N\sqrt{MT})$. Plugging this in (\ref{cumRewardReciprocalInequalityBandit}), we get the following:
\begin{align}
    \mathbb{E}\left[\dfrac{1}{R_i^{\alpha}(T)}\right] &\le O\left(N^2\left[\dfrac{1}{T^{\alpha}} + \dfrac{\hat{\text{Regret}}_T}{T}\right]\right) \nonumber\\
    &= O\left(N^3\sqrt{M}\left[\dfrac{1}{T^{\alpha}} + \dfrac{\log T}{\sqrt{T}}\right]\right) \nonumber\\
    &\overset{(a)}{=} O\left(N^3\sqrt{M}\left(\dfrac{1}{T^{\alpha}} + \dfrac{T^{\epsilon_0}}{\sqrt{T}}\right)\right)\nonumber\\
    &= O\left(\dfrac{N^3\sqrt{M}}{T^{\min(\alpha, \frac{1}{2} - \epsilon_0)}}\right)
\end{align}
where above in $(a)$, we have used the simple fact that $\log T = O(T^{\epsilon_0})$. Plugging the above bound in (\ref{reciprocalSquareInequalityBandit}), we get the following bound for any $0\le \alpha\le \frac{1}{2} - \epsilon_0$:
\begin{align}
    \hat{\text{Regret}}_T &\le \tilde{O}\left(\sqrt{MN}\sqrt{\sum_{t = 1}^T\sum_{i\in[N]}\mathbb{E}\frac{1}{R_i^\alpha(t - 1)}}\right) \nonumber\\
    &\le \tilde{O}\left(M^{\frac{1}{2} + \frac{1}{4}}N^{\frac{1}{2} + \frac{3}{2}}\sqrt{N\sum_{t = 1}^T \dfrac{1}{(t - 1)^\alpha}}\right) \nonumber\\
    &= \tilde{O}(M^{\frac{1}{2} + \frac{1}{4}}N^{\frac{1}{2} + \frac{3}{2} + \frac{1}{2}}T^{\frac{1 - \alpha}{2}}) \nonumber\\
    \label{baseCase1} &= \tilde{O}(M^{\frac{3}{4} }N^{\frac{5}{2}}T^{\frac{1 - \alpha}{2}})
\end{align}
By the same inequalities as above, for any $\frac{1}{2}  - \epsilon_0\le \alpha < 1$ we will have the bound:
\begin{align}
    \hat{\text{Regret}}_T &\le \tilde{O}\left(M^{\frac{3}{4}}N^{\frac{5}{2}}T^{\frac{1 - (\frac{1}{2} - \epsilon_0)}{2}}\right) \nonumber\\
    \label{baseCase2}&= \tilde{O}(M^{\frac{3}{4}}N^{\frac{5}{2}}T^{\frac{1}{4} + \frac{\epsilon_0}{2}})
\end{align}
More generally, suppose for some $0\le n < N_0$, we have
\begin{align}
    \label{inductionIneq1}\hat{\text{Regret}}_T &\le \tilde{O}(M^{\frac{2^{n + 2} - 1}{2^{n + 2}}}N^{\frac{2^{n + 3} - 3}{2^{n + 1}}}T^{\frac{1 - \alpha}{2}})\\
    \label{inductionIneq2}\hat{\text{Regret}}_T &\le \tilde{O}(M^{\frac{2^{n + 2} - 1}{2^{n + 2}}}N^{\frac{2^{n + 3} - 3}{2^{n + 1}}}T^{\frac{1}{2^{n + 2}} + \frac{2^{n + 1} - 1}{2^{n + 1}}\epsilon_0})
\end{align}
where (\ref{inductionIneq1}) holds for all $\alpha\in\left[\frac{2^n - 1}{2^n}, \frac{2^{n + 1} - 1}{2^{n + 1}} - \left(\frac{2^{n + 1} - 1}{2^n}\right)\epsilon_0\right]$, and (\ref{inductionIneq2}) holds for all $\alpha\in\left[\frac{2^{n + 1} - 1}{2^{n + 1}} - \left(\frac{2^{n + 1} - 1}{2^n}\right)\epsilon_0, 1\right)$. Note that, by our choice of $\epsilon_0$, both these intervals have non-negative measure (recall (\ref{epsilonIneq1})). Also, note that we have shown the base case for $n = 0$ via inequalities (\ref{baseCase1}) and (\ref{baseCase2}). We will now show that (\ref{inductionIneq1}) and (\ref{inductionIneq2}) continue to hold for $n + 1$.

Now, since we know that (\ref{inductionIneq2}) holds for all $\alpha\in\left[\frac{2^{n + 1} - 1}{2^{n + 1}} - \left(\frac{2^{n + 1} - 1}{2^n}\right)\epsilon_0, 1\right)$, we plug the bound (\ref{inductionIneq2}) in (\ref{cumRewardReciprocalInequalityBandit}) and get the following for such $\alpha$:
\begin{align}
    &\mathbb{E}\left[\dfrac{1}{R_i^\alpha(t)}\right] \nonumber\\
    &\le O\left(M^{\frac{2^{n + 2} - 1}{2^{n + 2}}}N^{\frac{2^{n + 3} - 3}{2^{n + 1}} + 2}\left[\dfrac{1}{T^\alpha} + \dfrac{\log T\cdot T^{\frac{1}{2^{n + 2}} + \frac{2^{n + 1} - 1}{2^{n + 1}}\epsilon_0}}{T}\right]\right) \nonumber\\
    &= O\left(M^{\frac{2^{n + 2} - 1}{2^{n + 2}}}N^{\frac{2^{n + 3} - 3}{2^{n + 1}} + 2}\left[\dfrac{1}{T^\alpha} + \dfrac{\log T}{T^{1 - \frac{1}{2^{n + 2}} - \left(\frac{2^{n + 1} - 1}{2^{n + 1}}\right)\epsilon_0}}\right]\right) \nonumber\\
    &\overset{(a)}{=} O\left(M^{\frac{2^{n + 2} - 1}{2^{n + 2}}}N^{\frac{2^{n + 3} - 3}{2^{n + 1}} + 2}\left[\dfrac{1}{T^\alpha} + \dfrac{T^{\epsilon_0}}{T^{1 - \frac{1}{2^{n + 2}} - \left(\frac{2^{n + 1} - 1}{2^{n + 1}}\right)\epsilon_0}}\right]\right) \nonumber\\
    &= O\left(M^{\frac{2^{n + 2} - 1}{2^{n + 2}}}N^{\frac{2^{n + 3} - 3}{2^{n + 1}} + 2}\left[\dfrac{1}{T^\alpha} + \dfrac{1}{T^{\frac{2^{n + 2} - 1}{2^{n + 2}} - \left(\frac{2^{n + 2} - 1}{2^{n + 1}}\right)\epsilon_0}}\right]\right) \nonumber\\
    \label{inductionStepIneq}&= O\left(M^{\frac{2^{n + 2} - 1}{2^{n + 2}}}N^{\frac{2^{n + 3} - 3}{2^{n + 1}} + 2}\left[\dfrac{1}{T^{\min \left(\alpha, \frac{2^{n + 2} - 1}{2^{n + 2}} - \left(\frac{2^{n + 2} - 1}{2^{n + 1}}\right)\epsilon_0\right)}}\right]\right)
\end{align}
where in $(a)$ above, we have simply used the fact that $\log T = O(T^{\epsilon_0})$. Note that by our choice of $\epsilon_0$ (recall inequality (\ref{epsilonIneq1})), we have
\begin{align}
    \dfrac{2^{n + 1} - 1}{2^{n + 1}}\le \dfrac{2^{n + 2} - 1}{2^{n + 2}} - \left(\dfrac{2^{n + 2} - 1}{2^{n + 1}}\right)\epsilon_0
\end{align}
So, plugging the bound of (\ref{inductionStepIneq}) in (\ref{reciprocalSquareInequalityBandit}), we can obtain
\begin{align}
    \label{inductionIneq3}\hat{\text{Regret}}_T &\le \tilde{O}(M^{\frac{2^{n + 3} - 1}{2^{n + 3}}}N^{\frac{2^{n + 4} - 3}{2^{n + 2}}}T^{\frac{1 - \alpha}{2}})\\
    \label{inductionIneq4}\hat{\text{Regret}}_T &\le \tilde{O}(M^{\frac{2^{n + 3} - 1}{2^{n + 3}}}N^{\frac{2^{n + 4} - 3}{2^{n + 2}}}T^{\frac{1}{2^{n + 3}} + \frac{2^{n + 2} - 1}{2^{n + 2}}\epsilon_0})
\end{align}
where (\ref{inductionIneq3}) holds for all $\alpha\in \left[\frac{2^{n + 1} - 1}{2^{n + 1}}, \frac{2^{n + 2} - 1}{2^{n + 2}} - \left(\frac{2^{n + 2} - 1}{2^{n + 1}}\right)\epsilon_0\right]$ and (\ref{inductionIneq4}) holds for all $\alpha\in \left[\frac{2^{n + 2} - 1}{2^{n + 2}} - \left(\frac{2^{n + 2} - 1}{2^{n + 1}}\right)\epsilon_0, 1\right)$. Hence, by induction, we see that (\ref{inductionIneq1}) and (\ref{inductionIneq2}) hold for all $0\le n\le N_0$ in the respective intervals.

Finally, (\ref{alphaKnotChoice}) and (\ref{epsilonIneq2}) imply that $\alpha_0\in \left[\frac{2^{N_0} - 1}{2^{N_0}}, \frac{2^{N_0 + 1} - 1}{2^{N_0 + 1}} - \left(\frac{2^{N_0 + 1} - 1}{2^{N_0}}\right)\epsilon_0\right]$. So, by what we've shown above, we conclude that 
\begin{align}
    \hat{\text{Regret}}_T\le \tilde{O}(M^{\frac{2^{N_0 + 2} - 1}{2^{N_0 + 2}}}N^{\frac{2^{N_0 + 3} - 3}{2^{N_0 + 1}}}T^{\frac{1 - \alpha_0}{2}})
\end{align}
and this completes the proof of the claim.

\subsection{Proof of \textbf{Lemma} \ref{puttaRegretBound}}\label{puttaBoundProof}

Fix some context $j\in[M]$ and a time horizon $T$. Consider the sequence $(\bm{g}_t)_{t:c_t = j}$ of all reward vector that context $j$ sees. Recall that $\bm{g}_t = \phi'(\bm{R}(t - 1))\odot \bm{r}(t)$. By our assumption, the reward vectors $\bm{r}(t)$ are generated by an \textit{oblivious adversary}, i.e they are fixed beforehand. However, the cumulative reward vectors $\bm{R}(t - 1)$ for $t\in[1, T]$ are \textit{policy-dependent}, i.e they are \textit{random}. Also, note that at every time step $t$, our policy picks some arm $I_t\in[N]$ to be played; from equation (\ref{cumRewardsBandit}) that defines how cumulative rewards are updated, we see that there are only finitely many sequences $(g_t)_{t:c_t = j}$ that our policy can see over the time horizon $T$. 

So, let $S$ be the set of all sequences $(\bm{g}_t)_{t:c_t = j}$ that our policy can see. Let $\bm{q}\in\Delta_S$ be the probability distribution that the policy induces over the set $S$ of possible reward sequences. For a fixed reward sequence $(\bm{l}_t)_{t:c_t = t}\in S$, we have the following by \textbf{Theorem} \ref{puttaOriginalBound}:
\begin{align}
    &\mathbb{E}\left[\max_{\{\bm{e}_k\}_{k = 1}^N}\sum_{t:c_t = j}\langle \bm{l}_t, \bm{e}_k - \bm{X}^j(t)\rangle\right] \nonumber\\
    &\le \tilde{O}\left(\sqrt{N\sum_{t:c_t = j}\lVert\bm{l}_t\rVert_2^2} + \max_{t:c_t = j}\lVert \bm{l}_t\rVert_\infty\sqrt{N\sum_{t:c_t = j}\lVert\bm{l}_t\rVert_1}\right)
\end{align}
Above, the expectation in the first time is taken w.r.t the policy actions. Now, taking expectations in the above inequality w.r.t the distribution $\bm{q}$ over $S$, we get
\begin{align}
    &\mathbb{E}\left[\mathbb{E}\left[\max_{\{\bm{e}_k\}_{k = 1}^N}\sum_{t:c_t = j}\langle \bm{l}_t, \bm{e}_k - \bm{X}^j(t)\rangle\right]\right] \nonumber\\
    \label{conditionalExpectationBound}&\le \mathbb{E}\left[\tilde{O}\left(\sqrt{N\sum_{t:c_t = j}\lVert\bm{l}_t\rVert_2^2} + \max_{t:c_t = j}\lVert \bm{l}_t\rVert_\infty\sqrt{N\sum_{t:c_t = j}\lVert\bm{l}_t\rVert_1}\right)\right]
\end{align}
By the tower property of conditional expectations, the first term on the LHS in the above inequality is just
\begin{align}
    &\mathbb{E}\left[\mathbb{E}\left[\max_{\{\bm{e}_k\}_{k = 1}^N}\sum_{t:c_t = j}\langle \bm{l}_t, \bm{e}_k - \bm{X}^j(t)\rangle\right]\right] \nonumber\\
    &= \mathbb{E}\left[\max_{\{\bm{e}_k\}_{k = 1}^N}\sum_{t:c_t = j}\langle \bm{g}_t, \bm{e}_k - \bm{X}^j(t)\rangle\right]
\end{align}
where the expectation on the RHS above is taken w.r.t the policy actions. Combining the above with (\ref{conditionalExpectationBound}), the claim follows.

\section{Additional Experiments} \label{addl_expts}

\begin{figure*}[!htb]
\minipage{0.45\textwidth}
  \includegraphics[width=0.75\linewidth]{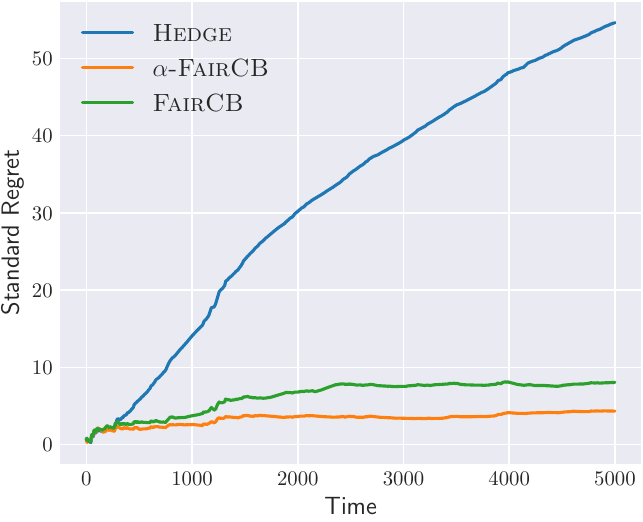}
  \caption{\scriptsize{Standard regret for the full information setting.}}\label{fig:standard_regret_full_info}
\endminipage\hfill
\minipage{0.45\textwidth}
  \includegraphics[width=0.75\linewidth]{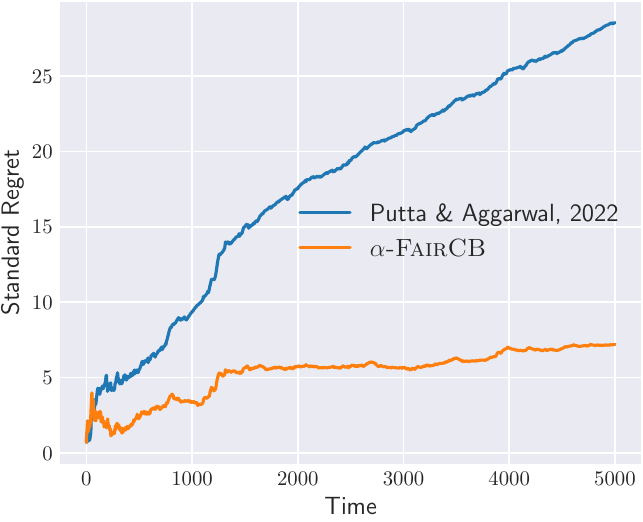}
  \caption{\scriptsize{Standard regret for the bandit information setting.}}\label{fig:standard_regret_bandit_info}
\endminipage\hfill
\end{figure*}

\begin{figure*}[!htb]
\minipage{0.45\textwidth}
  \includegraphics[width=0.85\linewidth]{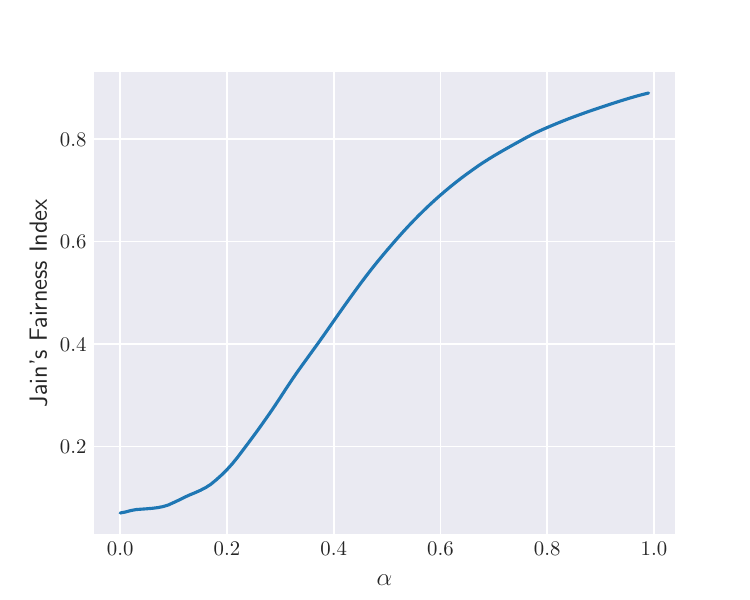}
  \caption{\scriptsize{Jain's Fairness Index for varying values of $\alpha$}}\label{fig:fairness_variation}
\endminipage\hfill
\minipage{0.45\textwidth}
  \includegraphics[width=0.85\linewidth]{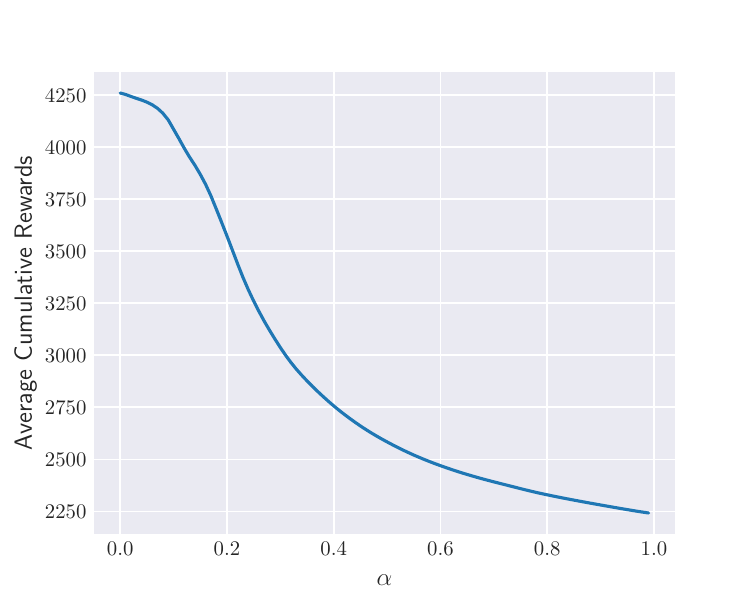}
  \caption{\scriptsize{Averaged Cumulative Rewards for varying values of $\alpha$}}\label{fig:avg_cumulative_rewards}
\endminipage\hfill
\end{figure*}

Figures \ref{fig:standard_regret_full_info} and \ref{fig:standard_regret_bandit_info} show plots of the standard regret of all the policies in the full information and the bandit information settings respectively. As before, it is clearly seen that the $\alpha\textsc{-FairCB}$ policy beats all the other policies in terms of the standard regret in both the settings.
 
We also run a few experiments to see how varying values of $\alpha$ in the interval $[0, 1)$ affects fairness levels in terms of Jain's Fairness Index (\ref{jain-fairness}). Note that the offline benchmark defined by equations (\ref{alphaFairRegretDefn}) and (\ref{contextualalphaPseudoRegret}) becomes the usual sum of rewards benchmark in the case of $\alpha = 0$, i.e it corresponds to an unfair objective. Thus, increasing values of $\alpha$ in the range $[0, 1)$ should increase fairness levels. Consequently, the average of cumulative rewards, defined by the equation
\begin{align}
    \label{avg_cumulative_rewards}
    \text{Average Cumulative Reward} := \dfrac{\sum_{i\in[N]} R_i(T)}{N}
\end{align}
should also decrease as $\alpha$ increases.

For these experiments, we only consider those contexts (users) in the dataset that occur with a high frequency (at least $5000$). In the dataset, there are $M = 18$ such contexts, and as before there are $N = 19$ arms (genres). In this case, the time horizon was $T = 136267$. We again sort the rows of the filtered dataset by timestamps. We took $100$ distinct values of $\alpha$ in the interval $[0, 1)$, and trained the $\alpha\textsc{-FairCB}$ policy for each of these values of $\alpha$ in the full information setting. Figure \ref{fig:fairness_variation} shows a plot of Jain's Fairness Index achieved by the $\alpha\textsc{-FairCB}$ policy for varying values of $\alpha$ (the index was calculated for the cumulative rewards at the final time step $T$). It is clearly seen that as $\alpha$ increases, the fairness index also increases. Figure \ref{fig:avg_cumulative_rewards} shows the average cumulative reward (\ref{avg_cumulative_rewards}, computed at the final time step $T$) and it is observed that as $\alpha$ increases, the average cumulative reward decreases.